\documentclass[10pt,twocolumn,letterpaper]{article}

\usepackage[pagenumbers]{wacv} 

\usepackage{graphicx}
\usepackage{amsmath}
\usepackage{amssymb}
\usepackage{booktabs}
\usepackage{comment}
\usepackage{adjustbox}

\usepackage{color}
\usepackage{epsfig}
\usepackage[normalem]{ulem}
\usepackage{algorithm}
\usepackage[noend]{algpseudocode}

\usepackage{pifont}
\usepackage{multirow}
\usepackage{xspace}

\usepackage{pgfkeys}
\usepackage{cite}

\usepackage{caption}
\usepackage{subcaption}

\usepackage[symbol]{footmisc} 

\usepackage[pagebackref,breaklinks,colorlinks]{hyperref}

\usepackage[capitalize]{cleveref}
\crefname{section}{Sec.}{Secs.}
\Crefname{section}{Section}{Sections}
\Crefname{table}{Table}{Tables}
\crefname{table}{Tab.}{Tabs.}

\begin{document}
\def\wacvPaperID{1548} 
\def\confName{WACV}
\def\confYear{2024}

\title{Preserving Image Properties Through Initializations in Diffusion Models}

\author{Jeffrey Zhang\\
{\tt\small jeff@revery.ai}
\and
Shao-Yu Chang\\
{\tt\small shaoyuc3@illinois.edu}
\and
Kedan Li\\
{\tt\small kedan@revery.ai}
\and
David Forsyth\\
{\tt\small daf@illinois.edu}
}
\maketitle

\begin{abstract}
Retail photography imposes specific requirements on images.  For instance, images may need uniform background colors, consistent model poses, centered products, and consistent lighting.  Minor deviations from these standards impact a site's aesthetic appeal, making the images unsuitable for use.  We show that Stable Diffusion methods, as currently applied, do not respect these requirements. The usual practice of training the denoiser with a very noisy image and starting inference with a sample of pure noise leads to inconsistent generated images during inference. This inconsistency occurs because it is easy to tell the difference between samples of the training and inference distributions. As a result, a network trained with centered retail product images with uniform backgrounds generates images with erratic backgrounds. The problem is easily fixed by initializing inference with samples from an approximation of noisy images. However, in using such an approximation, the joint distribution of text and noisy image at inference time still slightly differs from that at training time. This discrepancy is corrected by training the network with samples from the approximate noisy image distribution. Extensive experiments on real application data show significant qualitative and quantitative improvements in performance from adopting these procedures. Finally, our procedure can interact well with other control-based methods to further enhance the controllability of diffusion-based methods.
\end{abstract}

\section{Introduction}
\label{sec:intro}

Stable Diffusion~\cite{rombach2021highresolution} can generate high-quality, lifelike images, and has opened up numerous
innovative applications. Examples include creating new art, style transfer between
pictures, and generating high-resolution images from text. However many real applications require
images to meet specific design requirements.  For example, product images need consistent
photographic standards so that retail websites maintain uniform aesthetic appeal and are ``on-brand''.
Even minor deviations can make images unusable.

The essential requirements for our application mirror those of many other applications.  Our application requires a
text-to-image generator where:
\textbf{(1) outputs reflect a garment description text accurately};
\textbf{(2) outputs are either a garment image or a human model wearing the described garment};
\textbf{(3) garments are not cropped, are centered, and appear on a white background};
\textbf{(4) human models are always depicted from foot to neck, stand in similar poses, and appear on a neutral background};
\textbf{(5) outputs have consistent professional lighting  and shading};
and \textbf{(6) one text-to-image model can produce all desired images and does not produce others}.
The first five requirements ensure images are ``on-brand'' and the last is for efficiency. Remarkably, Stable Diffusion~\cite{rombach2021highresolution} as currently practiced cannot meet these requirements, but quite simple changes result in a model that does.

We start from the observation that fine-tuning Stable Diffusion~\cite{rombach2021highresolution} with product images on neutral backgrounds {\em does not} produce a method that can generate product images on neutral backgrounds (Fig.~\ref{fig:dataset}). This unexpected effect is caused by a hiccup in the structure of the method.  Current training methods \cite{song2020denoising, NEURIPS2020_4c5bcfec} form a weighted sum of noise and a base image (using a weight $\alpha$), and a model is trained to denoise the noisy image. At inference, one assumes that a sufficiently noisy base image is indistinguishable from noise and that the denoising process can be started with pure noise. We show the assumption is true only for $\alpha$ much smaller than those used in current practice, meaning that the denoiser sees noticeably different distributions at train and test times. Training Stable Diffusion~\cite{rombach2021highresolution} for very small $\alpha$ is also challenging (see Supplementary). 

A simple alternative is to initialize with a draw from a distribution that represents the training distribution reasonably well and is easily sampled.  While training, some information about the original image can be recovered from the initial noisy sample, which at best is a blurry version of the original image. This means that a heavily noised sample from a mixture of principal components model is an acceptable approximation of the training distribution.
We show significant improvement results from using this as an initial distribution {\em without} retraining Stable Diffusion~\cite{rombach2021highresolution} -- for instance, erratic backgrounds are replaced by neutral backgrounds (see Supplementary).   
But these improvements highlight another initialization problem: the {\em joint} distribution between text and noisy image at training is misrepresented by both standard noise initialization and our initialization. We show that because our initialization is easily sampled, it can be used in fine-tuning the denoiser, leading to notable improvements in text-based control.
Finally, we show that our initialization techniques are easily integrated with other controllability methods (e.g. ControlNet~\cite{zhang2023adding}) to provide more effective control for diffusion-based methods. 

\section{Related work}
There is considerable research available on generating specific concepts or subjects using diffusion-based methods. However, to the best of our knowledge, we believe that we are the first to concentrate on creating specific image distributions for images.

\subsection{Object preservation and harmonization}
There are works that learn specific objects and generate variations of those objects faithfully. Dreambooth~\cite{ruiz2022dreambooth} fine-tunes a pretrained diffusion model to accurately generate new variations of a particular subject. Gal \etal~\cite{gal2022textual} invert objects into pseudo-words to attain personalized text embeddings to create images of those objects. Other works have enabled editing in the forward pass of diffusion models without image-specific fine-tuning or inversion. 
InstructPix2Pix~\cite{brooks2022instructpix2pix} takes an input image and generates a new image based on text instructions. Yang \etal~\cite{PaintByExample} proposes an exemplar-based image editing model where the reference image is semantically transformed and harmonized into another image. Finally, Edward \etal~\cite{hu2022lora} adapts the language models to generate specific objects by adding trainable parameters of the language embeddings to learn new concepts from a dataset. This allows the adaptation of new words and concepts by fine-tuning the newly added parameters instead of the entire generation model.

These studies primarily focus on preserving target objects and generally struggle to control non-target areas if no conditions exist in those areas. Unlike these approaches, our paper emphasizes stabilizing the image distributions throughout the diffusion process. Our proposed method can effectively preserve the properties of the entire image, not just the target objects.

\subsection{User-defined controllablility}
Many of the works mentioned above are text-guided, in which users provide a text prompt to control and edit images. However, language-guided manipulations often do not generate images satisfying users' requirements. CLIP features~\cite{CLIP} leverage the representations of user-provided images to improve the diversity of the output results. Region-based image editing methods~\cite{DiffEdit, Avrahami_2022_CVPR} treat the task as a conditional inpainting task with a mask highlighting the regions of the images that needed to be edited while preserving non-target areas. To enhance task-specific control of diffusion models, ControlNet~\cite{zhang2023adding} adds an additional input condition (e.g. edge maps, segmentation masks, keypoints, etc.) alongside text prompts to manipulate image generation. 

While these methods allow users to control target areas in the images by adding additional conditions, they still often fail to maintain image distributions. This failure is due to inconsistencies in the initialization process during training and inference. In Sec.~\ref{sec:controlNet_application}, we show that combining our method with ControlNet~\cite{zhang2023adding} can strengthen the control abilities of diffusion models and stabilize the output (Fig.~\ref{fig:warp_initialization}).

\subsection{Image-to-Image Translation}
Finally, Stable Diffusion~\cite{rombach2021highresolution} is often applied to image-to-image translation applications by choosing a starting image and generating variations from the image initialization. The resulting images have significant similarities to the initial image's colors and contours. On the other hand, other methods use DDIM Inversion~\cite{song2020denoising} to find initial noise vectors to restore the original image during diffusion to apply image-to-image translation. In Tune-A-Video~\cite{wu2022tuneavideo}, the authors use DDIM Inversion to control the consistency of frames and the contours of objects. In Null-text Inversion~\cite{mokady2022null}, DDIM Inversion is used to create images similar in appearance to the original input, enabling users to edit specific words while preserving the objects of the original image.

In these works, reference images are used to generate variations of that reference image. In contrast, we demonstrate a sample from an approximate noisy image distribution significantly changes the behavior of diffusion-based methods because the method experiences similar samples from the training distribution at inference time. Furthermore, we show that using the right starting initialization during both training and inference is essential for consistently generating entire image distributions, not just for maintaining the features of a specific reference image.

\section{Method}
\label{sec:method}

Current literature on Stable Diffusion~\cite{rombach2021highresolution} has assumptions at inference time that appear inconsequential, but we show these assumptions have significant consequences. We demonstrate substantial improvements in resolving these errors at inference time. 

\subsection{Background}
Stable Diffusion~\cite{rombach2021highresolution} is trained to recover an image $x_0$ by denoising a noisy image $x_t$ at timestep $t \in [0,T]$. At the $t$'th timestep, the denoiser is presented with
\begin{equation} \label{eq:x_t}
    x_t = \sqrt{\alpha_t} x_0 + \sqrt{(1-\alpha_t)}\epsilon_t
\end{equation}
where $\epsilon_t \sim \mathcal{N}(0,I)$ and $\alpha_t$ is the cumulative product of scaling at each timestep $t$ (refer to DDIM~\cite{song2020denoising}). Following the training procedure from~\cite{song2020denoising}, a denoiser $f$ predicts the added noise $\hat{\epsilon}_t = f(x_t, t, e; \Theta)$, where $f$ is parameterized by $\Theta$ and takes in noisy image $x_t$, timestep $t$, and conditional encoding $e$. We write $\hat{x}_{0}$ for the predicted ground truth image derived from removing the predicted noise $\hat{\epsilon}_t$ from $x_t$. Our loss is  
\begin{equation} \label{eq:loss}
    \mathcal{L} = \mathbb{E}[||\epsilon_t - \hat{\epsilon}_t||_2^2]
\end{equation}

From Eq.~\ref{eq:x_t}, if we have the predicted $\hat{\epsilon}_t$ and the noisy image $x_t$, we can derive the predicted ground truth image $\hat{x}_0$ with
\begin{equation} \label{eq:x_0}
    \hat{x}_0 = (x_t - \sqrt{(1-\alpha_t)}\epsilon_t)/ \sqrt{\alpha_t} 
\end{equation}

\begin{figure}[t!]
    \centering
    \begin{subfigure}[t]{0.5\textwidth}
        \centering
         \includegraphics[width=\textwidth,trim=4 4 4 4,clip]{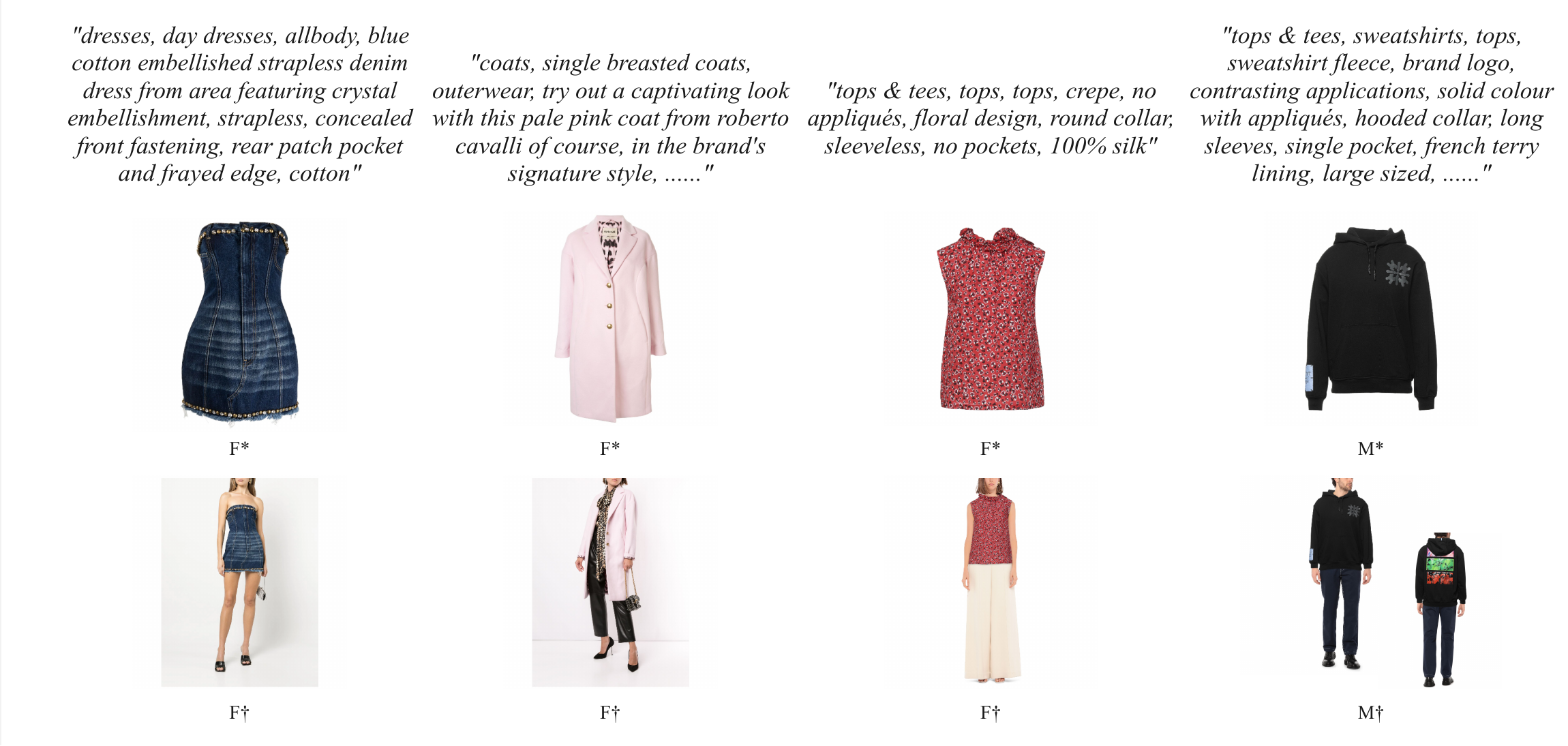}
         
         \caption{Garment + Model + Text Dataset}
         \label{fig:dataset(a)}
    \end{subfigure}%
    
    \begin{subfigure}[t]{0.5\textwidth}
        \centering
         \includegraphics[width=\textwidth,trim=4 4 4 4,clip]{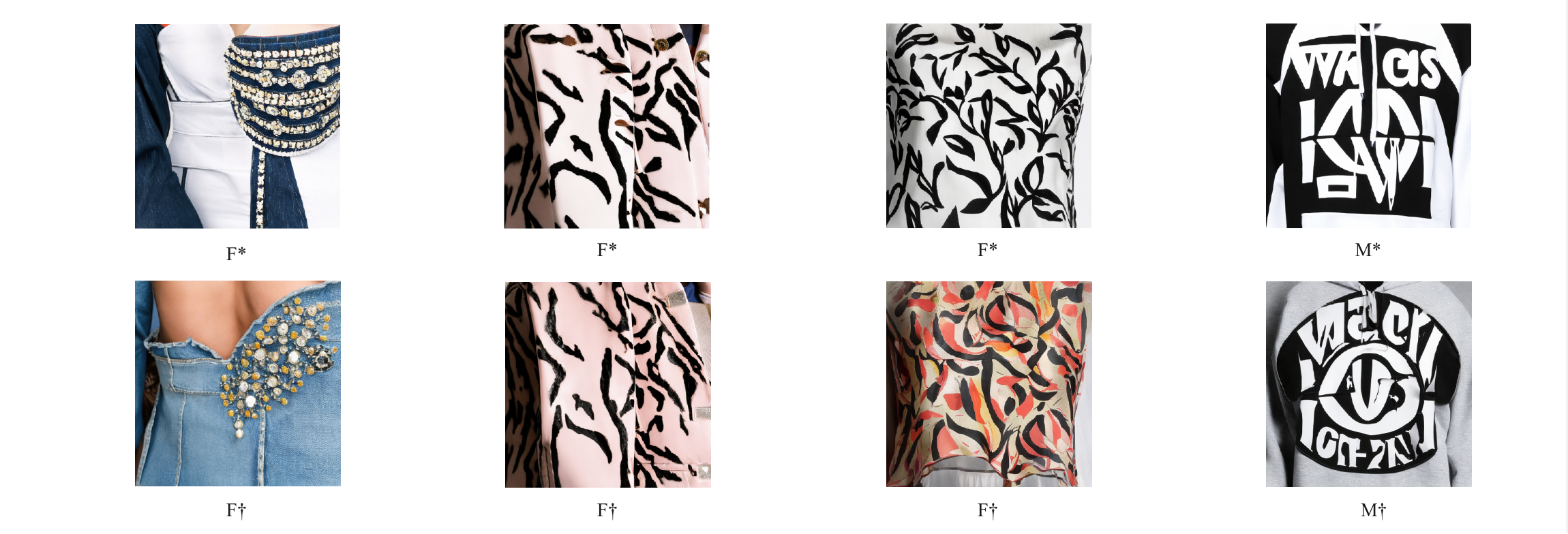}
         \caption{DDIM Training + Inference}
         \label{fig:dataset(b)}
    \end{subfigure}
    
    \begin{subfigure}[t]{0.5\textwidth}
        \centering
         \includegraphics[width=\textwidth,trim=4 4 4 4,clip]{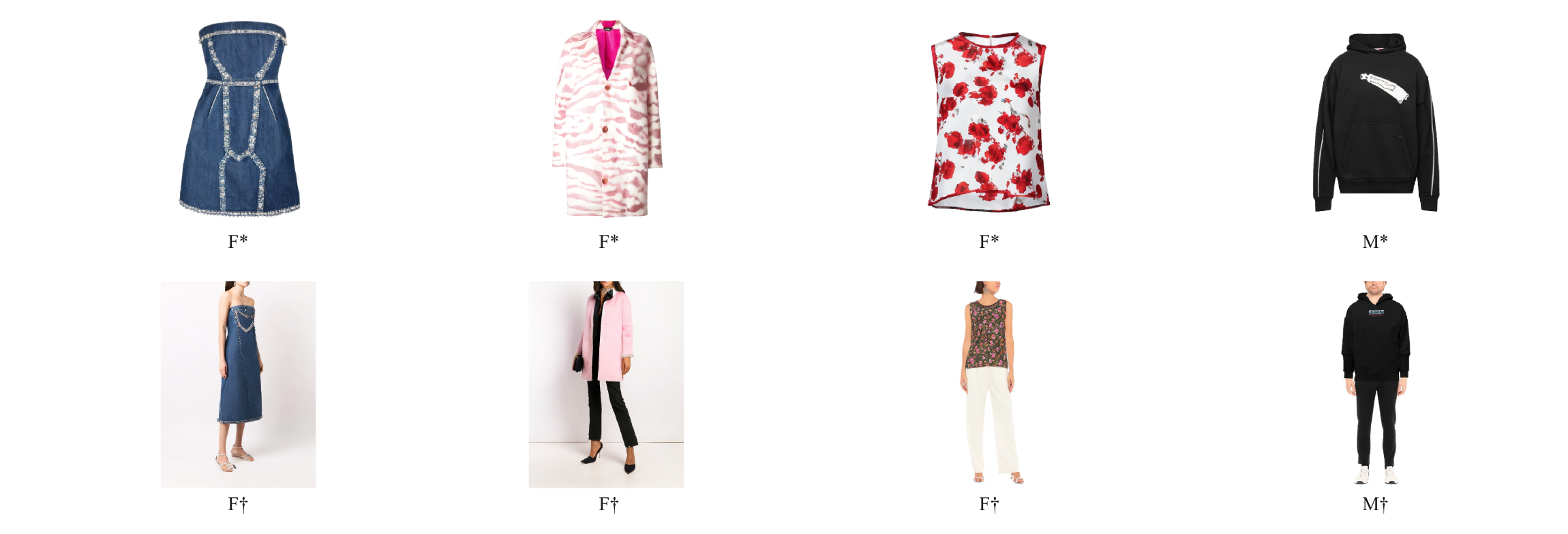}
         \caption{DDIM Training + Control Inference ($x_{start}=x_0$)}
         \label{fig:dataset(c)}
    \end{subfigure}
    \caption{Despite training on images with properties (1)-(5), normal diffusion-based training and inference lead to unexpected results. (a) shows sample sequences from our garment dataset. (b) shows standard fine-tuning and inference results with Stable Diffusion~\cite{rombach2021highresolution} do not generate the same distribution of images despite being trained on images from (a). The prompts are taken from training data, where we expect the best results. To show that this is not a training error, in (c), we set a control experiment by changing $x_{start}$ (Eq. \ref{eq:pca_offset_inference}) to the training image shown in (a). The generated images match the training distribution, indicating that initialization information strongly influences results. (F*: "female garment, no person, white background"; M*: "male garment, no person, white background"; F†: "female person wearing garment"; M†: "male person wearing garment")}
    \label{fig:dataset}
    \vspace{-0.05in}
\end{figure}

\subsection{Inference Assumption}
During training, we have ground truth image $x_0 \sim P(images)$ and initial noisy image $x_T \sim P_T$ (from Eq.~\ref{eq:x_t}). However, at inference time, we do not have the ground truth image $x_0$ and must supply an alternative initialization $x_{init}$.
Following \cite{song2020denoising}, it is usual to argue that for sufficiently small $\alpha_T$, $x_T$ should be very similar to $\mathcal{N}(0,I)$. Hence, during inference, a common approach is to sample our initialization $x_{init} \sim \mathcal{N}(0,I)$.


However, if $\alpha_T$ is not small enough, then $x_T$ has information about $x_0$ that the network could recover and utilize for denoising. Pure noise as initialization may not behave as expected for two reasons: (1) denoising networks are trained on noticeably different data distributions than what they see at inference time and (2) denoising networks may extract information about $x_0$ from $x_T$ to denoise $x_T$. If this is true, reliable inference procedures might use something other than $\mathcal{N}(0,I)$.

We show values of $\alpha_T$ in many pretrained models may indeed be too large. In Fig.~\ref{fig:dataset}, we compare the performance of a Stable Diffusion~\cite{rombach2021highresolution} fine-tuned to make images of models and garments on two different initializations. In Fig.~\ref{fig:dataset(b)}, we initialize $x_{init} \sim \mathcal{N}(0,I)$ during inference, and in Fig.~\ref{fig:dataset(c)}, we initialize with $x_{init} = x_T$ during inference using Eq.~\ref{eq:x_t}, where $x_0$ is the ground truth image from Fig.~\ref{fig:dataset(a)}. Different initializations have significantly different results, but more importantly, notice the network takes obvious hints from the ground truth image $x_0$ (in Fig.~\ref{fig:dataset(c)}).




For values of $\alpha_T$ that are not small enough, it is possible to reliably distinguish between samples from $P_T$ and $\mathcal{N}(0,I)$ using elementary methods. For $x_{init} \sim \mathcal{N}(0,I)$ to be hard to distinguishable from $x_T \sim P_T$, we must have
\begin{equation} \label{eq:alpha_small}
    \alpha_T < O(\frac{1}{d}),
\end{equation}
where $d = H \times W \times C$ is the dimension of the sampled image. Meeting this constraint is difficult as $1/d$ is significantly smaller than current values of $\alpha_T$. The key point is that spatial averages of images have strong properties and will be perceptible even at small $\alpha_T$.

Set $\textbf{a} = \frac{1}{d}\textbf{1}$ where $\textbf{1}$ is the 1-vector of size $d$. Then, we can write
\begin{equation} \label{eq:expected_value}
    \mathbb{E}_{P_T}[\textbf{a} \cdot x_T] = \sqrt{\alpha_T} \mathbb{E}_{P(images)} [\textbf{a} \cdot x_0] = \sqrt{\alpha_T}\mu \neq 0,
\end{equation}
where $\mu = \mathbb{E}_{P(images)}[\textbf{a} \cdot x_0]$ is the average of images. Simple experiments show that $\mu$ is non-zero, and different sets of images can have different values of $\mu$ (e.g., white background, people in similar poses, etc.).
From Eq.~\ref{eq:expected_value}, $\textbf{a} \cdot x_T$ has mean  $\mu_T = \sqrt{\alpha_T}\mu$
and variance $\sigma_{T}^2 = \alpha_T \sigma^2 + \frac{(1 - \alpha_T)}{d}$.
But for $x_{init} \sim \mathcal{N}(0,I)$, $\textbf{a} \cdot x_{init}$ has mean 0 and $\sigma^2 = 1/d$. 

Hence, the network is presented with samples from two distributions, $\mathcal{N}(\mu_T, \sigma^2_T)$ during training and $\mathcal{N}(0, \sigma^2)$ during inference. For the distributions to be difficult to distinguish, we want $\mu_T/\sigma$ and $\mu_T/\sigma_T$ to be small. We have
\begin{equation} \label{eq:uT_sigma}
    (\frac{\mu_T}{\sigma})^2 = d\alpha_T(\mu)^2,
\end{equation}
and
\begin{equation} \label{eq:uT_sigmaT}
    \small
    (\frac{\mu_T}{\sigma_T})^2 = \frac{(\sqrt{\alpha_T}\mu)^2}{\alpha_T\sigma^2 + (1-\alpha_T)\frac{1}{d}} = \frac{d\alpha_T\mu^2}{\alpha_T(\sigma^2d-1)+1}
\end{equation}

Consequently, both Eq.~\ref{eq:uT_sigma} and~\ref{eq:uT_sigmaT} are small if $\alpha_T$ is less than $1/d$ or smaller, giving us Eq.~\ref{eq:alpha_small}.

Practical numbers are $d = 64\times64\times4 = 16384$ and $\alpha_T = 0.0047$; but $0.0047 \nless 0.000061$, hence $\alpha_T$ is not in the right range (Eq.~\ref{eq:alpha_small}). This indicates that a denoiser network $f$ could tell the difference between the initialization samples used in training and those used at inference. If it can tell the difference, different behaviors between training and inference are possible. Fig.~\ref{fig:dataset} demonstrates the network actually behaves differently for samples with these different distributions. 


One solution is to train with a smaller $\alpha_T$. However, training with smaller $\alpha_T$ requires scaling down $\alpha_t$ for many timesteps, which leads to more difficult training as more noise is added (see Supplementary). We show that approximating $P(images)$ offers a more efficient and reliable solution.

\subsection{PCA-K Offset Inference}
\label{sec:proposed_mean_offset}

Our procedure \textbf{``PCA-K Offset Inference"} initializes inference with:
\begin{equation} \label{eq:proposed_inference}
    x_{init} = \sqrt{\alpha_T} x_{start} + \sqrt{1-\alpha_T}\epsilon_T,
\end{equation}
where $x_{start}$ is sampled from a distribution $Q$ that approximates $P(images)$. 
$Q$ does not need to be a particularly strong approximation of $P(images)$ because a large magnitude of noise is added in $x_T$. Because this noise is i.i.d. Gaussian noise, we expect that the information the network can extract about $x_{start}$ from $x_{init}$ is a heavily smoothed version of $x_{start}$.  
Hence, we need a distribution model that is easy to sample, reasonably approximates blurry images, and can handle multiple classes. 



PCA-K Offset Inference uses a mixture of normals for $Q$, where each normal is derived from Principal Component Analysis (PCA) of images of its class $c$. PCA is known to be an effective description of blurred images. $K$ represents the number of principal components. For each class $c$ in our dataset, we model an image as 
\begin{equation} \label{eq:pca_offset}
x_R^c = \mu^c + \sum_{i=1}^{K} \xi_i \textbf{p}_i
\end{equation}
where $\xi_i \sim N(0, \lambda_i)$, $\textbf{p}_i$ are orthonormal principal components, and $\mu_c$ is the mean image of class $c$. Write $x_R$ for a random image drawn from an R principal component model. Then setting $x_{start} = x_R^c$, our initialization becomes:
\begin{equation} \label{eq:pca_offset_inference}
    x_{init} = \sqrt{\alpha_T} x_R^c + \sqrt{1-\alpha_T}\epsilon_T
\end{equation}
We call this very useful case where $R=0$, $x_{start}$ is the class mean $\mu^c$, \textbf{``Mean Offset Inference"}. 

\subsection{PCA-K Offset Training}

While PCA-K Offset Inference allows the inference procedure to mimic the training procedure much more closely, we still expect operating conditions to differ. Our approximation may not exactly match the distribution used in training. In particular, our approximation may not preserve the delicate relationship between the ground truth image $x_0$ and the text encoding $e$ supplied during training. 
Fig.~\ref{fig:dataset} shows improvements by using an approximate distribution during inference time that was trained as usual, but further improvements are available. At training, the network sees a noisy version of an image and models a complex relationship between image and text encoding. But at inference, our network will be presented with a text encoding $e$ and a sample $x_{start}$ from the initial distribution, which has not been conditioned on $e$. We cannot guarantee an approximate distribution can preserve those relationships. We can, however, use the same approximate distribution during training so that the network experiences the same distribution at train time and test time. This is a significant advantage of a $Q$ that is easy to sample.

We use $x_{init}$ from Eq.~\ref{eq:proposed_inference} in place of the initialization $x_T$ in Eq.~\ref{eq:x_t} to resemble the desired start point for image $x_0$. This allows the network to train and infer from the same distribution for the first step of the diffusion process:
\begin{equation} \label{eq:mean_offset_training}
    x_{T}^{new} = \sqrt{\alpha_T} x_{start} + \sqrt{1-\alpha_T}\epsilon_T
\end{equation}
First, we want the denoiser to recover the ground truth image $x_0$ from $x_{T}^{new}$. So, we alter the noise objective to  
\begin{equation} \label{eq:epsilon_T_new}
    \epsilon_{T}^{new} = (x_{T}^{new} - x_0 \sqrt{\alpha_T})/ (\sqrt{1-\alpha_T})
\end{equation}
Second, we want to skip timesteps (for computation efficiency, as in DDIM's~\cite{song2020denoising}), so this change must be applied to multiple timesteps. Let $S$ be the number of skips per timestep. Then, we apply the mean offset training to all timesteps within the first skip step to guarantee the first skip step is trained with the mean offset initialization. Hence, we alter Eq.~\ref{eq:x_t} and Eq.~\ref{eq:epsilon_T_new} for $t \ge T - S$ to 
\begin{equation} \label{eq:x_t_new}
    x_{t}^{new} = \sqrt{\alpha_t} x_{start} + \sqrt{1-\alpha_t}\epsilon_t
\end{equation}
and 
\begin{equation} \label{eq:epsilon_new}
    \epsilon_{t}^{new} = (x_{t}^{new} - x_0 \sqrt{\alpha_t})/ (\sqrt{1-\alpha_t})
\end{equation}
Thus, our final loss is a combination of Eq.~\ref{eq:loss} and the new noise objective from Eq.~\ref{eq:epsilon_new}:
\begin{equation} \label{eq:new_loss}
\mathcal{L}_{new} = 
\begin{cases}
        \mathbb{E}[||\epsilon_t - \hat{\epsilon}_t||_2^2] & \text{if } t < T - S\\
        \mathbb{E}[||\epsilon_{t}^{new} - \hat{\epsilon}_t||_2^2] & \text{if } t \ge T-S
\end{cases}
\end{equation}

During training, we project a ground truth image $x_0$ with class $c$ into $x_K^c$ and set $x_{start} = x_K^c$ from Eq.~\ref{eq:pca_offset}, giving our proposed \textbf{``PCA-K Offset Training"} procedure. Setting $K=0$ gives us $x_K = \mu^c$, which is simply just initializing $x_{start} = \mu^c$ and is our \textbf{``Mean Offset Training"} procedure. We find Mean Offset Training works best compared to higher $K$ values and is much simpler in practice (see Supplementary). As a result, we use Mean Offset Training for results in the main text for the sake of simplicity.



\section{Experiments and Results}
\label{sec:results}
In these experiments, we use data collected from retailers that follow properties specified in Sec.~\ref{sec:intro} (details in Sec.~\ref{sec:dataset}). We show PCA-K Offset Inference behaves better because the initialization is similar to the training distribution in the initial denoising timesteps (Sec.~\ref{sec:naive_results}). We show sampling a bad initialization from PCA-K can damage the relationship between text and initialization during inference because the operating conditions are still different (Sec.~\ref{sec:naive_results} and Fig.~\ref{fig:init_bias}). We show that incorporating our PCA-K Offset Training procedure fixes this issue (Sec.~\ref{sec:offset_training_results} and Fig.~\ref{fig:mean_initialization}). Finally, we show other control methods experience the same initialization problem, and our procedures can be easily combined with other methods to provide further control in generation (Sec.~\ref{sec:controlNet_application}). 

\subsection{Dataset}
\label{sec:dataset}
We collect over a million image pairs of retailer garment, garment on model, and garment text description triplets (Fig.~\ref{fig:dataset(a)}). We are given one text prompt corresponding to a garment image and a model wearing that garment. All training data triplets satisfy \textbf{properties (1)-(5)} described in Sec.~\ref{sec:intro}. Our task is to generate images of garments and fashion models wearing garments that satisfy all properties (1)-(6). To distinguish between generating garments and models wearing garments, we prepend the caption with "male/female garment, no person, white background" (denoted with \textbf{M*} and \textbf{F*}, respectively) for generating garments and "male/female person wearing garment" for generating human models (denoted with \textbf{M†} and \textbf{F†}, respectively). 

For inference, we collect 24 different freeform text descriptions of garments by asking fashion designers to describe diverse garment descriptions (see Supplementary). These are fictional garment descriptions used to test the generalizability of our text-to-image model.

\subsection{DDIM Finetuning with PCA-K Offset Inference}
\label{sec:naive_results}

\begin{figure}[t!]
    \centering
    \includegraphics[width=0.49\textwidth,trim=4 4 4 4,clip]{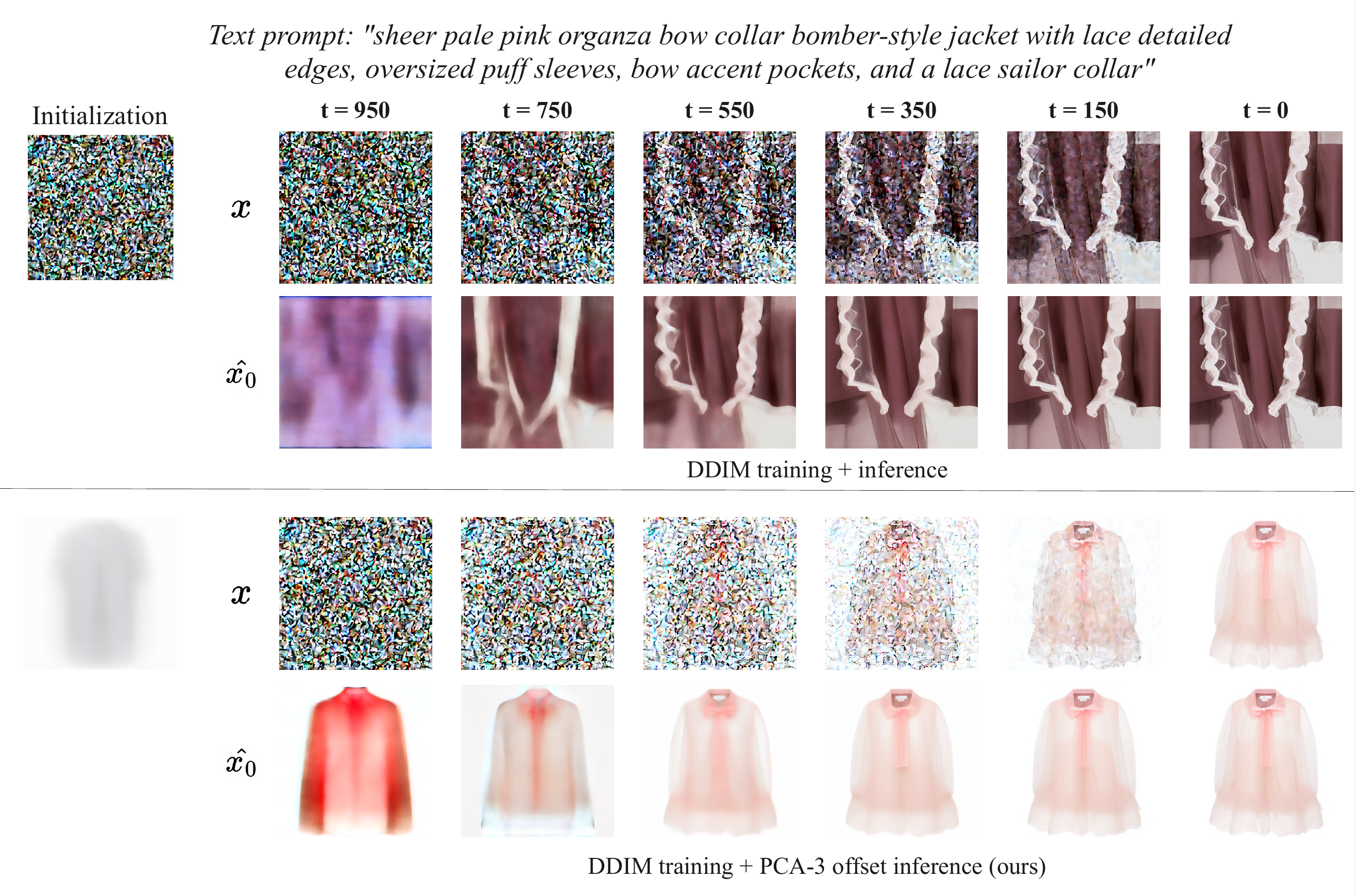}
    \caption{
    Intermediate outputs for a $S=20$ DDIM training process are visualized to show the first step of the diffusion process is out of distribution for standard DDIM training + inference. The top two rows show the intermediate outputs when initializing with noise (DDIM inference). The bottom two rows show the intermediate outputs when projecting a gray sweater with PCA-3 Offset Inference. Rows 1 and 3 show the noisy image $x_t$ and rows 2 and 4 show the predicted $\hat{x_0}_t$ at each time step $t$. We can see from row 2 that the first predicted $\hat{x_0}$ introduces a dark, non-uniform background that is propagated throughout the process, whereas in row 4, the predicted $\hat{x_0}$ is already close to the desired distribution, making the diffusion process is much more stable.}
    \label{fig:init_intermediate}
\end{figure}

\begin{figure*}[htp!]
    \centering
    \includegraphics[width=0.92\textwidth, trim=4 4 4 4,clip]{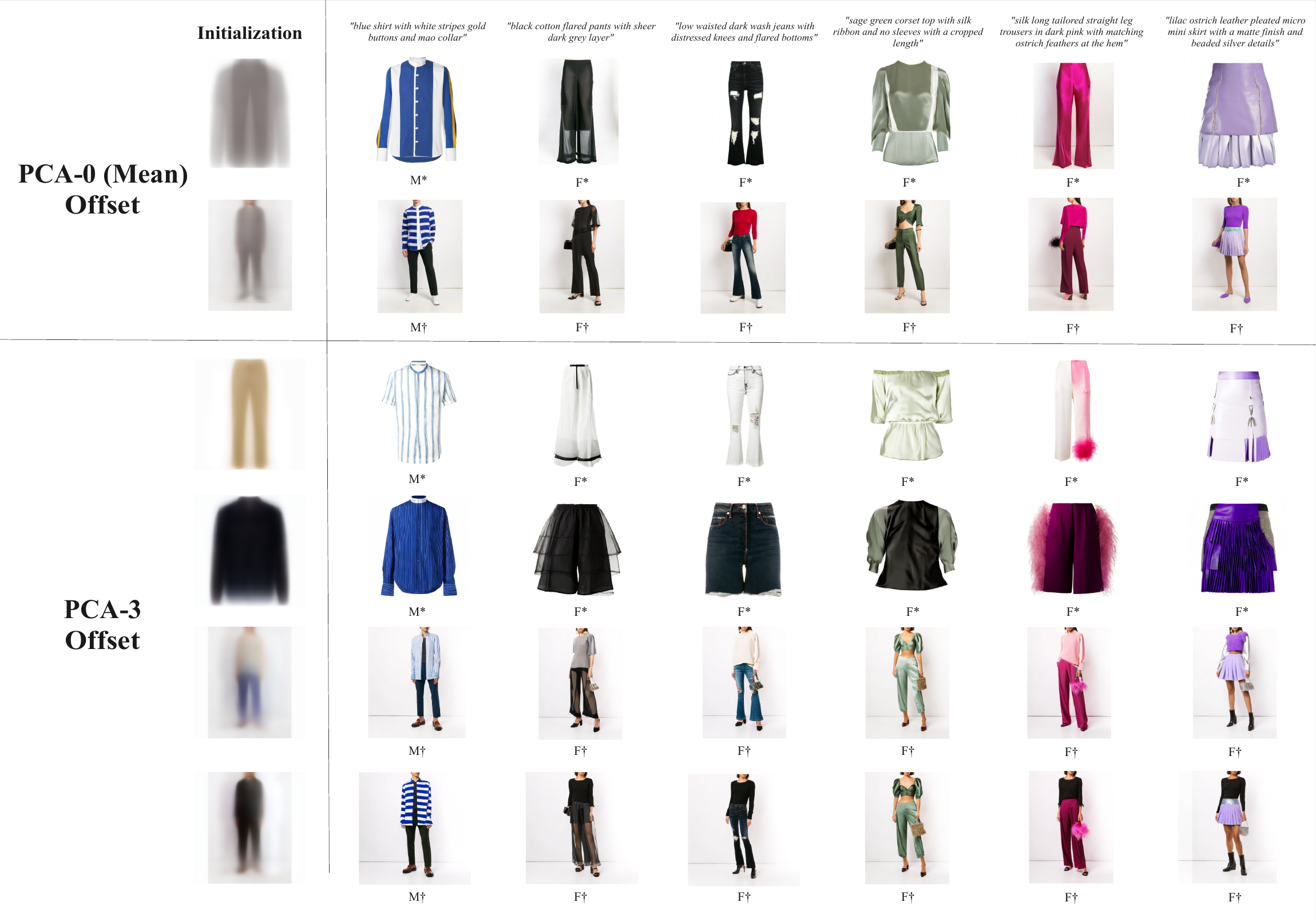}
    \caption{We show applying our PCA-0 and PCA-3 Offset Inference on DDIM Training can significantly improve generating desired image properties but is strongly biased by the sampled initialization $x_{start}$. This leads to some undesirable artifacts. Rows 1 and 2 show PCA-0 occasionally generates non-white backgrounds for garments due to faint sleeves in the mean image - violating \textbf{property (3)}. In rows 3-6, generated images are strongly influenced by the color and shape of $x_{start}$ and ``...black cotton flared pants..." are generated to be white, ``...tailored straight leg trousers..." are generated as shorts, etc. This violates (\textbf{property (1)}) as the generated images do not respect the text and further indicate that $x_{start}$ strongly influences the denoiser. Descriptions are freeform text from fashion designers.}
    \label{fig:init_bias}
\end{figure*}

To set a baseline, we fine-tune sd-v1.5~\cite{rombach2021highresolution} on our dataset and show results from different initializations in Fig.~\ref{fig:dataset}. We fine-tune for 50,000 steps on a batch size 16 and learning rate 1e-5. We use a set total number of timesteps $T=1000$ for training and skip timesteps $S=50$ for inference (i.e., 20 total timesteps for inference). We indicate the standard training and inference procedure from~\cite{song2020denoising} as \textbf{DDIM training} and \textbf{DDIM inference}, respectively.

\textbf{Different inference initializations have qualitatively different effects}. Fig.~\ref{fig:dataset(b)} shows that if we initialize with noise (DDIM inference), none of the generated images respect \textbf{properties (1)-(6)} despite being fine-tuned on images with those properties. These images are not curated; generated images hardly ever show isolated garments or models. Furthermore, the text prompts used for inference \textit{were taken from the training data}, where we expect the best behavior. To show that this is not a training bug, we set $x_{start}$ in Eq.~\ref{eq:proposed_inference} to the ground truth dataset images shown in Fig.~\ref{fig:dataset(a)}. The fine-tuned network can now generate images that satisfy all our desired image properties (Fig.~\ref{fig:dataset(c)}). This indicates that $x_{start}$ in the initialization heavily impacts the denoising process, and the assumption that $\mathcal{N}(0,I)$ is close enough to $P_T$ has clear implications during inference (using the actual ground truth is not the issue here; below and Fig.~\ref{fig:init_bias}). 

\textbf{Visualizing the mechanism by displaying intermediate time steps in generation helps to understand the impact of different initializations better}. In Fig.~\ref{fig:init_intermediate}, we compare the results between random noise initialization (DDIM training + inference) and initializing with a gray garment from our dataset projected to 3 PCA components (PCA-3 Offset Training + Inference) on freeform text from fashion designers. We see that during the first few steps of the diffusion process, a random noise input will predict a $x_0$ that is very different from our desired image distribution. This mistake is not corrected in later timesteps and is accumulated throughout the denoising process. This is because the denoiser is not trained to denoise an image from a non-white background. If we initialize with a PCA-projected garment image with a white background, then the training data distribution is maintained in all intermediate steps.

\textbf{PCA-K Inference fixes distribution issues, but initializations strongly affect text control in generations}. We apply PCA-K Inference on the fine-tuned model and show image distribution problems are mitigated in Fig.~\ref{fig:init_bias}, but the starting point can bias the type of images generated. Notice the initialization alters the shape and color of the garment generated because the network $f$ is trained to take hints from this initialization. The lighter initialization in row 3 generates more light colors, while row 4 shows much darker and boxier generations that adhere to the color and shape of the initialization. Notice that the generated garments do not respect the text ("black cotton flared pants..." are generated as white in the third column, "...straight leg trousers..." are generated as shorts in the sixth column, etc.), thus not satisfying \textbf{property (1)}. We try to apply a more neutral initialization by setting $x_{start}$ to garment and model means. However, Fig.~\ref{fig:init_bias} row 1 shows artifacts in the background due to the faint sleeve of the mean garment image. 

The denoiser clearly takes hints from the initialization when generating images. This further substantiates that $x_0$ has a tremendous weight during training. The denoiser heavily relies on cues from $x_0$ to denoise the image because $x_0$ strongly correlates with the encoding text $e$. Unfortunately, $x_{start}$ is sampled from a distribution $Q$ that is independent of the text prompt. As a result, the word control for denoising worsens when we sample from $Q$.

\begin{figure}[t!]
    \centering
    \begin{subfigure}[t]{0.49\textwidth}
        \centering
        \includegraphics[width=\textwidth,trim=4 4 4 4,clip]{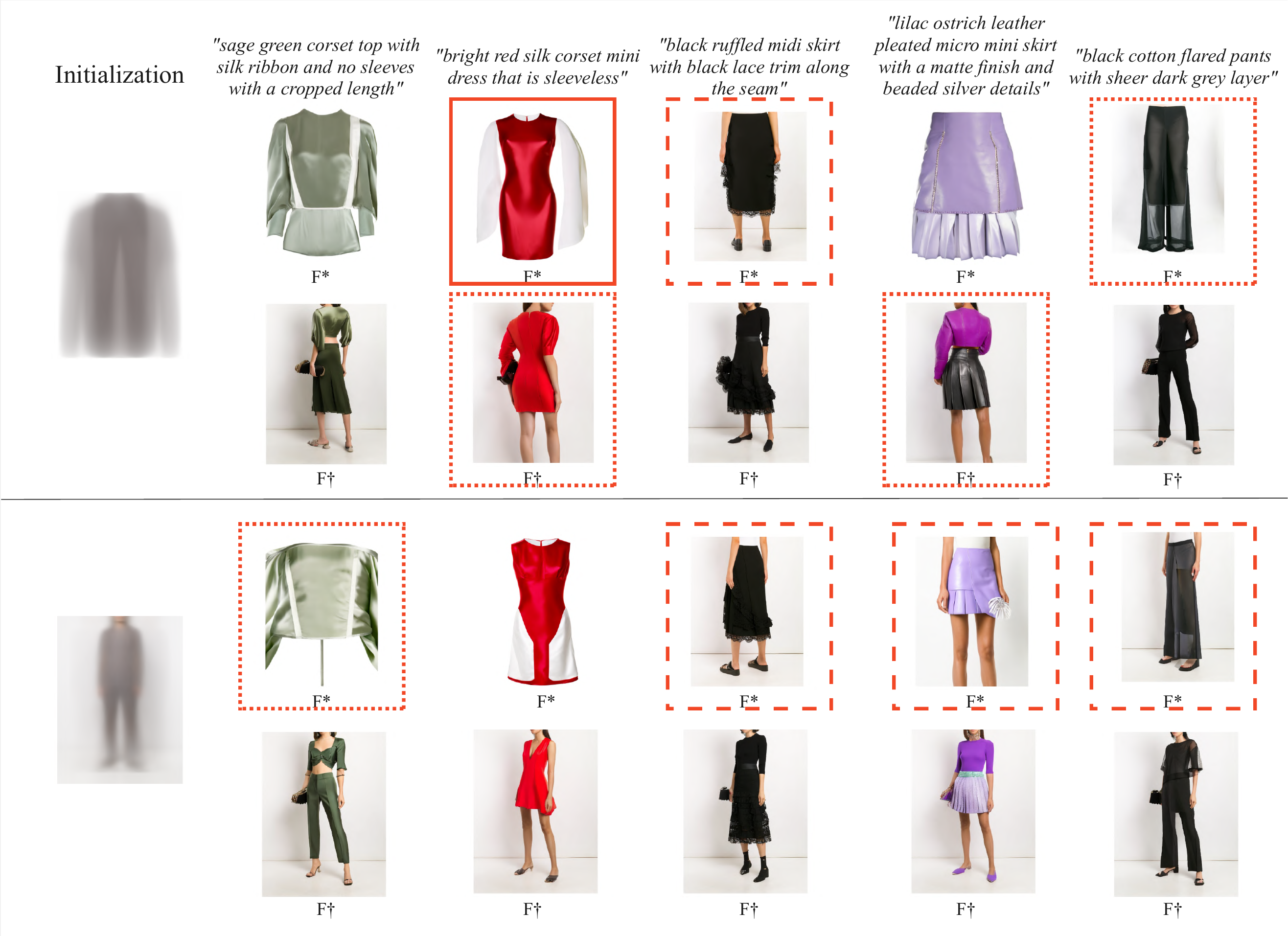}
        \caption{DDIM Training + Mean Offset Inference}
        \label{fig:mean_initialization_inference}
    \end{subfigure}%
    
    \begin{subfigure}[t]{0.49\textwidth}
        \centering
        \includegraphics[width=\textwidth,trim=4 4 4 4,clip]{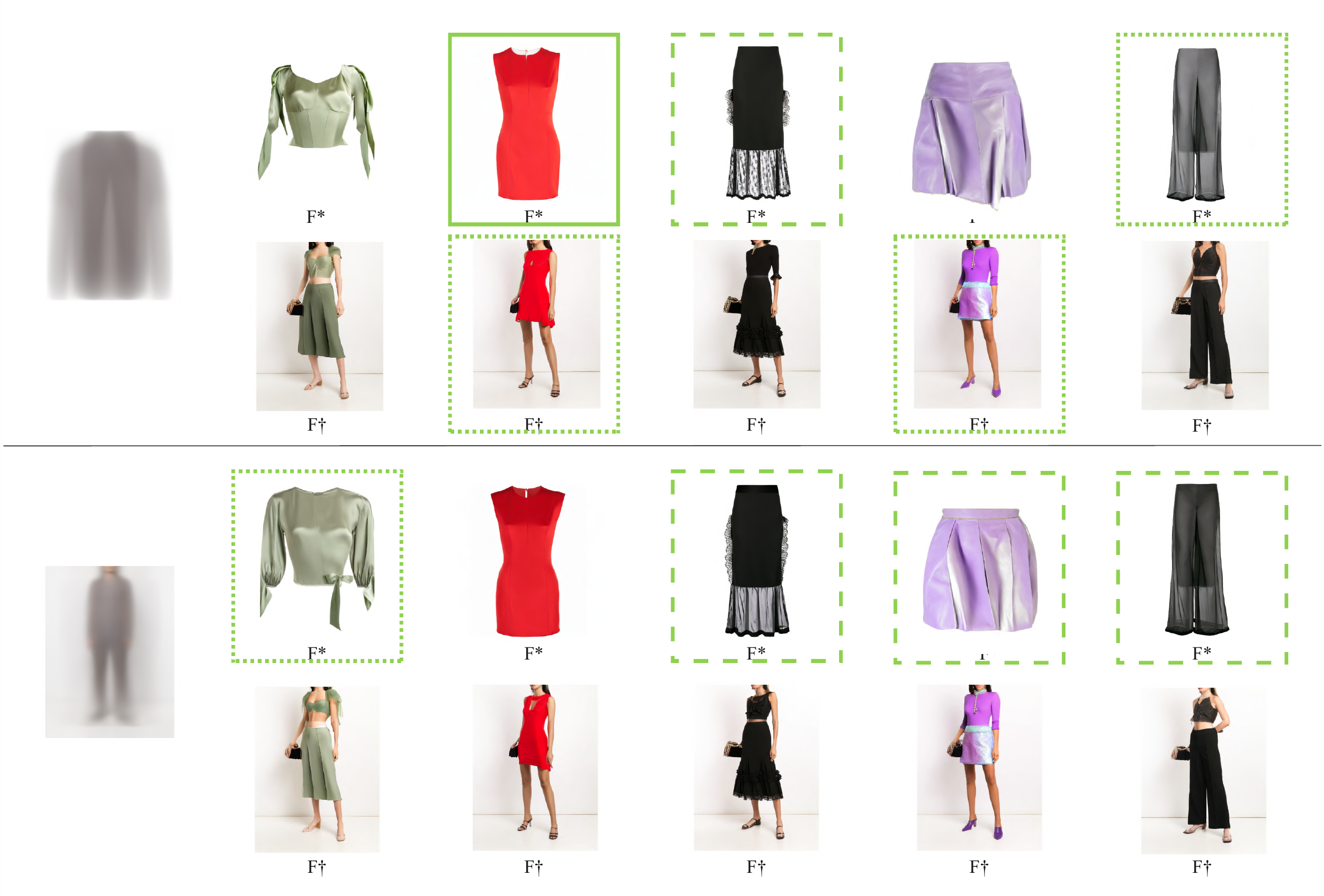}
        \caption{Mean Offset Training + Inference}
        \label{fig:mean_initialization_training}
    \end{subfigure}
    \caption{Using Mean Offset Training and Mean Offset Inference provides better text control because the relationship between initialization and text is preserved during training and inference. We apply two class mean initialization for garments and models and intentionally swap the means to test the effect of different initializations during inference. Figure (a) shows garment and model results DDIM Training + Mean Offset Inference that violate various properties. Figure (b) shows Mean Offset Training + Inference results satisfy all desired properties. Red boxes highlight generation errors in (a) and green boxes show they are fixed in (b). Red solid borders show artifacts that shouldn't exist and don't fully respect the text ((a) fails \textbf{property (1)}). Red dashed borders show generated models instead of garments, as specified by the text, and the person is not in the proper pose ((a) fails \textbf{properties (1) and (4)}). Red dotted borders show non-white backgrounds or cropped garments/models ((a) fails \textbf{property (3)}).}
    \label{fig:mean_initialization}
\end{figure}

\subsection{PCA-K Offset Training}
\label{sec:offset_training_results}
By incorporating PCA-K Offset Training, we alter the training procedure to have consistent initialization and text pairings during training. We test Mean Offset Training (PCA-0) with average garment and average model initializations. Training hyperparameters are identical to the fine-tuned model in Sec.~\ref{sec:naive_results}. Results for PCA-K ($K>0$) are shown in Supplementary.


\textbf{Qualitatively, Mean Offset Training generates images that respect the text better than standard training and satisfies all desired properties (1)-(6)}. Fig.~\ref{fig:mean_initialization} shows the difference between using DDIM training + Mean Offset Inference (Fig.~\ref{fig:mean_initialization_inference}) and Mean Offset Training + Inference (Fig.~\ref{fig:mean_initialization_training}). While only Mean Offset Inference significantly helps generate our desired image properties, it occasionally produces artifacts (not pure white backgrounds), does not accurately follow the text (generates sleeves when there shouldn't be sleeves), and crops the garments and models. Incorporating mean offset initialization in training resolves these issues and generates our desired image distribution (see results with all 24 text prompts in the Supplementary).

\begin{figure*}[t!]
     \centering
     \includegraphics[width=0.95\textwidth]{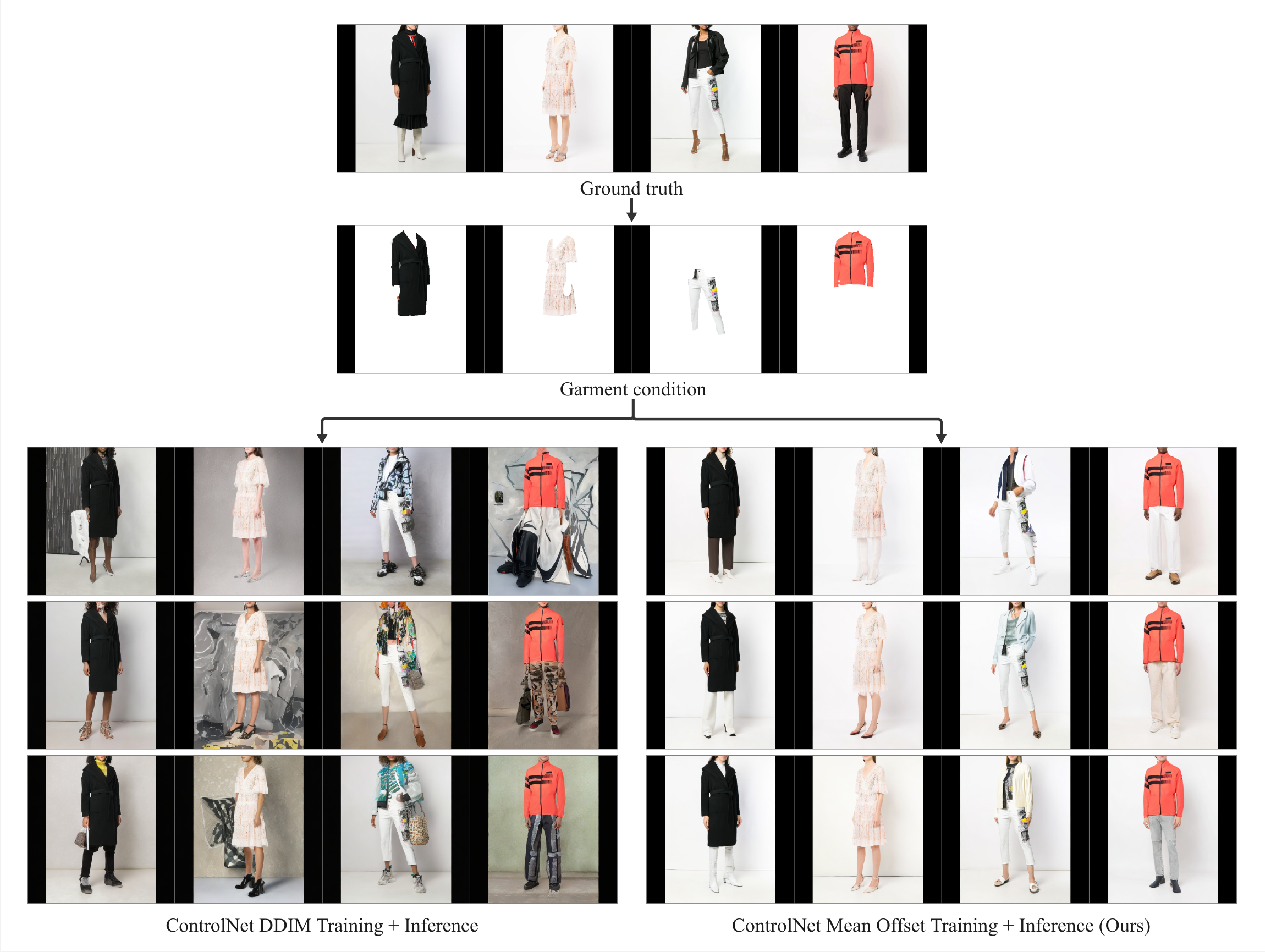}
     \label{fig:warp_initialization}
    \caption{We adapt ControlNet~\cite{zhang2023adding} to take a garment condition to generate models wearing garments. We display three seeds for the same control to show that vanilla ControlNet (DDIM Training + Inference) consistently produces out-of-distribution results (violating \textbf{properties (4) and (5))}, whereas ControlNet with Mean Offset Training + Inference (Ours) perfectly preserves the desired training distribution.}
    \label{fig:warp_initialization}
\end{figure*}

\textbf{Quantitatively, Mean Offset Training + Inference generates more accurate images} as demonstrated by running CLIP similarities~\cite{CLIP} between generated garments and text prompts in Table \ref{tab:clip_similarity}. Notice the $10.6\%$ improvement from DDIM Training + DDIM Inference to Mean Offset Training + Inference. While CLIP similarity is not a perfect representation of similarity, the improvement is significant. The learned text-to-initialization relationship is better preserved because the same initialization distribution is used during training and inference. 

\begin{table}
\centering
\resizebox{\columnwidth}{!}{
\begin{tabular}{ |c|c|c| }
    \hline
    Method & CLIP similarity($\uparrow$) & Score scaled by GT($\uparrow$)\\
    \hline
     Ground Truth Garments &  0.294 & 1.0\\
     \hline
      \hline
     DDIM Training + DDIM Inference & 0.2542 & 0.865\\
     \hline
     DDIM Training + Mean Offset Inference (Ours) & 0.2761 & 0.939\\
    \hline
     \textbf{Mean offset Training + Inference (Ours)} & \textbf{0.2812} & \textbf{0.956}\\
    \hline
\end{tabular}
}

\caption{We use CLIP similarity~\cite{CLIP} (higher is better) between images and text to show our methods generate images that respect text better (\textbf{property (1)}). We compare with the ground truth image from our dataset to set a baseline. Notice our Mean Offset Inference easily outperforms standard DDIM training and inference. Furthermore, incorporating DDIM Training + Inference further improves performance, indicating that a better text-to-generation relationship is preserved (for property (1)).}
\label{tab:clip_similarity}
\end{table}

\subsection{Application to ControlNet} \label{sec:controlNet_application}
We apply our method to ControlNet~\cite{zhang2023adding} for a different task of virtual try-on using our dataset. We show image distribution issues persist in this new task due to noise initialization during inference, but can be fixed with our PCA-K Offset Training + Inference. For this task, we are given a garment as control and train a denoiser to generate a realistic person wearing that garment. To train, we mask the region of a garment from a person in our dataset and adapt ControlNet~\cite{zhang2023adding} to take the masked garment image as the condition to generate the remaining image. The left column of Fig.~\ref{fig:warp_initialization} shows the effect of training and running ControlNet without any modification to the initialization procedure, and background and lighting properties are not preserved (\textbf{properties (4) and (5)}). The right column of Fig. \ref{fig:warp_initialization} shows that applying our Mean Offset Training + Inference preserves desired generated image properties. 


\section{Discussion}
We believe our approximate initialization distribution has broad applications, not just limited to fashion retail images. Situations where images follow structural requirements could benefit from our training and inference procedure. 
Additionally, because sampling initialization can bias the inference (Fig.~\ref{fig:init_bias}), we intend to investigate using CCA to build relationships between PCA-K initializations and text features. 

\section{Conclusion}
Our research indicates that existing training and inference procedures for diffusion-based methods are problematic and cannot preserve certain image distributions. We uncover that the assumption of employing random noise as the starting point may significantly affect the way images are generated. 
More importantly, we show the current training procedure is largely biased by its initialization, but can be mitigated by adopting our PCA-K Offset Training + Inference. 
Finally, we demonstrate that our work is orthogonal to other manipulation methods, such as ControlNet~\cite{zhang2023adding}, and can be combined to enable greater control of diffusion-based image generation.


    
    


\newpage
{\small
\bibliographystyle{ieee_fullname}
\bibliography{main}
}

\end{document}


\def\wacvPaperID{1548} 
\def\confName{WACV}
\def\confYear{2024}

\title{Preserving Image Properties Through Initializations in Diffusion Models - Supplementary Materials}

\author{Jeffrey Zhang\\
{\tt\small jeff@revery.ai}
\and
Shao-Yu Chang\\
{\tt\small shaoyuc3@illinois.edu}
\and
Kedan Li\\
{\tt\small kedan@revery.ai}
\and
David Forsyth\\
{\tt\small daf@illinois.edu}
}
\maketitle


\section{Freeform Text Test Set}
For our test dataset, we asked three fashion designers to provide eight freeform descriptions of garments to test the generalizability of our text-to-image model. These are completely fictional garment descriptions. The 24 freeform texts we used for testing are:
\begin{itemize}
    \setlength\itemsep{0em} 
    \item (F* or F†) sheer pale pink organza bow collar bomber-style jacket with lace detailed edges, oversized puff sleeves, bow accent pockets, and a lace sailor collar
    \item (F* or F†) short sheer white dress with white shirt collar
    \item (F* or F†) baggy blue jeans with red stitching and silver buttons
    \item (M* or M†) blue shirt with white stripes gold buttons and mao collar
    \item (F* or F†) asymmetrical blue velvet dress with crochet white pocket
    \item (F* or F†) white sleeveless dress with ocean blue lining at the bottom
    \item (F* or F†) black sleeveless dress with ocean blue lining at the bottom
    \item (F* or F†) black hoodie with white embroidered flowers sleeves
    \item (F* or F†) black cotton flared pants with sheer dark grey layer
    \item (F* or F†) low waisted dark wash jeans with distressed knees and flared bottoms
    \item (M* or M†) brown leather moto jacket with silver buttons and zipper and lace trim from the bottom of the jacket
    \item (F* or F†) sage green corset top with silk ribbon and no sleeves with a cropped length
    \item (F* or F†) black ruffled midi skirt with black lace trim along the seam
    \item (F* or F†) tailored dark wash denim zip corset with pronounced seams and obi inspired bows at the hip
    \item (F* or F†) silk long tailored straight leg trousers in dark pink with matching ostrich feathers at the hem
    \item (F* or F†) lilac ostrich leather pleated micro mini skirt with a matte finish and beaded silver details
    \item (F* or F†) mock black neck cashmere sleeveless top with a heart cutout at the chest and silver threads throughout
    \item (F* or F†) sheer turquoise silk gown with deep V neckline and train hem, with ruffles on the sleeves and neckline and a cinched tie waist belt
    \item (F* or F†) leopard print chiffon column gown with a mock neck halter closure, no sleeves, and symmetrical hip cut outs
    \item (F* or F†) royal blue velvet puffer jacket with an oversized silhouette, stand up scarf collar, silver hardware, and extra long extended sleeves
    \item (F* or F†) military jacket with clean cut tailoring in dark emerald brocade fabric with crimson red piping and gold accents
    \item (F* or F†) silver cyber edgy style metal-look top with sleek tribal design that wraps around the body and tie in back with a silver strap to keep it on
    \item (F* or F†) patchwork pieced together handkerchief skirt with a bohemian style that is maxi length and looks hand-sewn, in many different prints and shades of green fabrics
    \item (F* or F†) asymmetrical cut denim wrap skirt in a black faded wash. Silver hardware inspired by Chrome Hearts. Rough unfinished hem and one side is longer on the wrap
\end{itemize}

We use M*/F* to shorthand the description ``male/female garment, no person, white background" and we use M†/F† to shorthand the description ``male/female person wearing garment." Whenever we aim to generate a garment, we use either M* or F*; whenever we aim to generate a model, we use M† and F†. \textbf{All images for these text prompts are displayed in order from left to right and top to bottom, unless mentioned otherwise in the figure.}

\section{Lowering $\alpha$ Values}
Compared to standard training parameters, lowering $\alpha$ values improves the stability of certain properties, but outputs still exhibit inconsistencies in preserving image properties and are often lower quality and underexposed. 

We test different values of $\alpha_T$ by adjusting the linear end parameter. Stable Diffusion~\cite{rombach2021highresolution} uses a linear start of $0.00085$ and a linear end of $0.012$. We keep the same linear start for all experiments and adjust the linear end to lower the magnitude of $\alpha_T$. We keep all other parameters identical to our finetuning training procedure in the main text. Table~\ref{tab:linear_end_alphas} shows the altered linear end values and each corresponding $\alpha_T$ value. 

\begin{figure*}[htp!]
    \centering
    \includegraphics[width=0.85\textwidth, trim=4 4 4 4,clip]{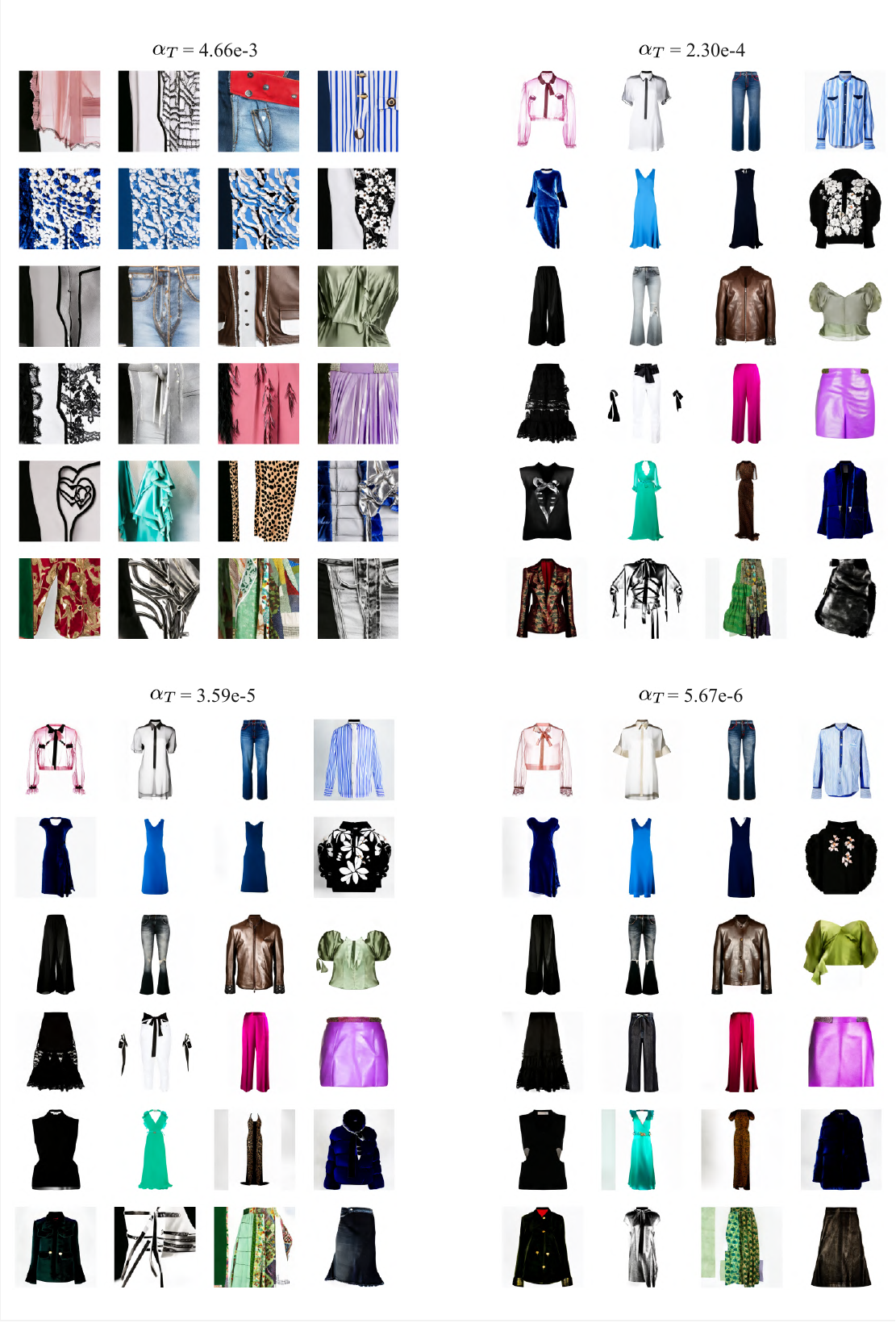}
    \caption{Comparison between fine-tuning on different $\alpha_T$ values for garment generation. We use standard noise initialization during inference.}
    \label{fig:garment_linear_end}
\end{figure*}

\begin{figure*}[htp!]
    \centering
    \includegraphics[width=0.85\textwidth, trim=4 4 4 4,clip]{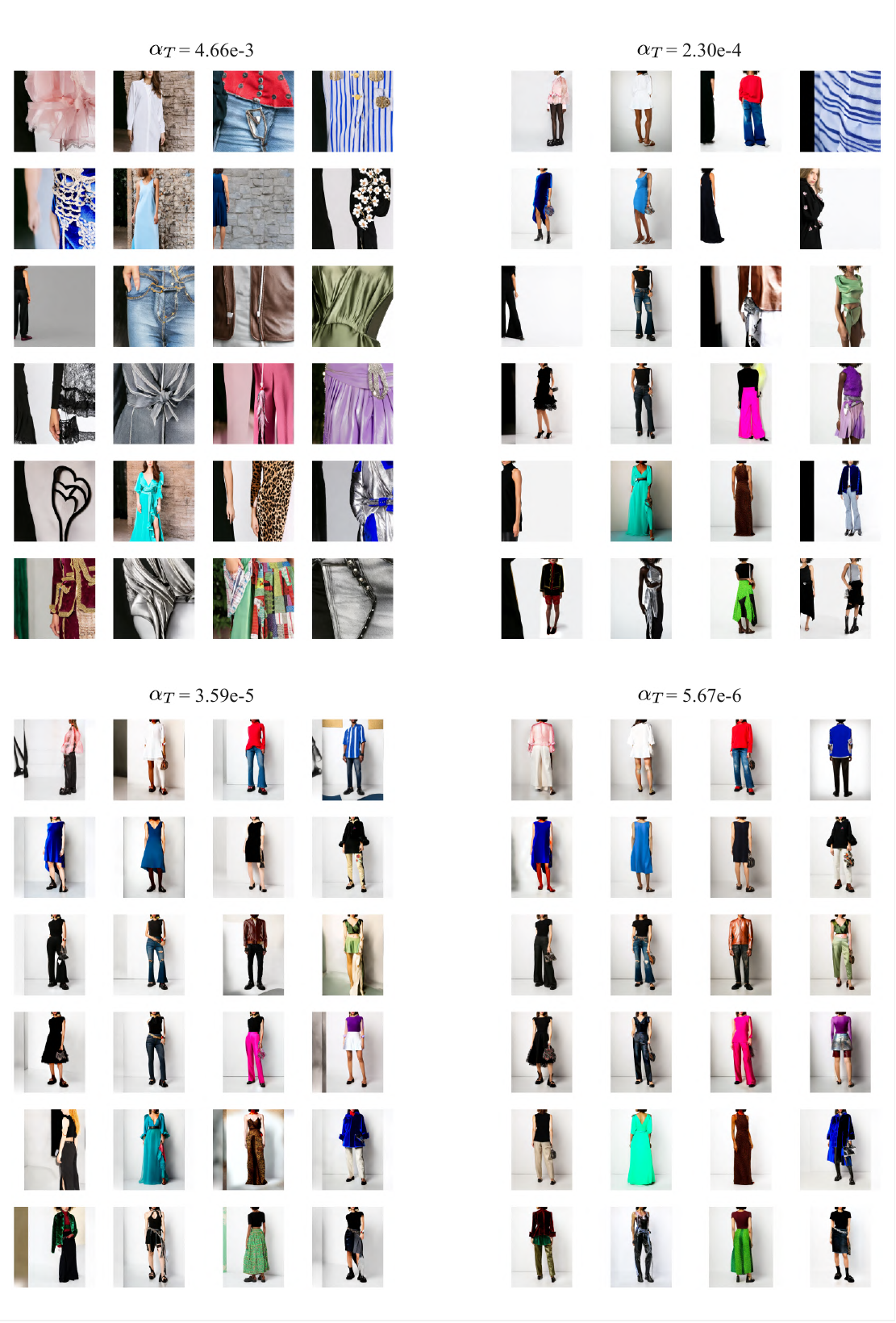}
    \caption{Comparison between fine-tuning on different $\alpha_T$ values for model generation. We use standard noise initialization during inference.}
    \label{fig:model_linear_end}
\end{figure*}

Fig.~\ref{fig:garment_linear_end} and Fig.~\ref{fig:model_linear_end} show results for garments and models from fine-tuning sd v1.5~\cite{rombach2021highresolution} on our dataset with different $\alpha_T$ values. Though decreasing $\alpha_T$ brings us closer to our preferred image properties, these properties aren't consistently produced, and the image quality is significantly compromised. 

\begin{figure*}[htp!]
    \centering
    \includegraphics[width=1.0\textwidth, trim=4 4 4 4,clip]{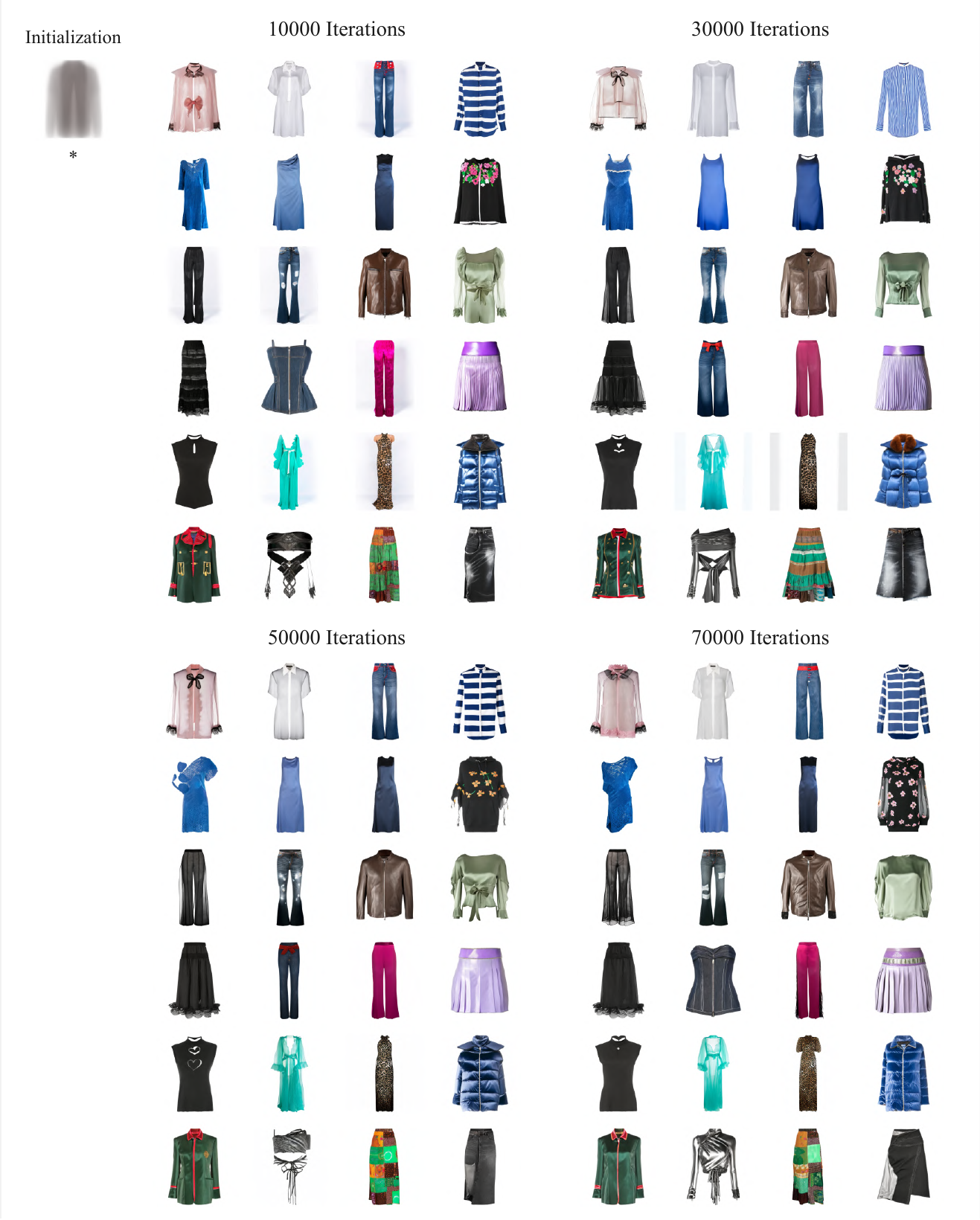}
    \caption{Comparison between training iterations on garment generation for Mean Offset Training + Inference.}
    \label{fig:garment_iteration}
\end{figure*}

\begin{figure*}[htp!]
    \centering
    \includegraphics[width=1.0\textwidth, trim=4 4 4 4,clip]{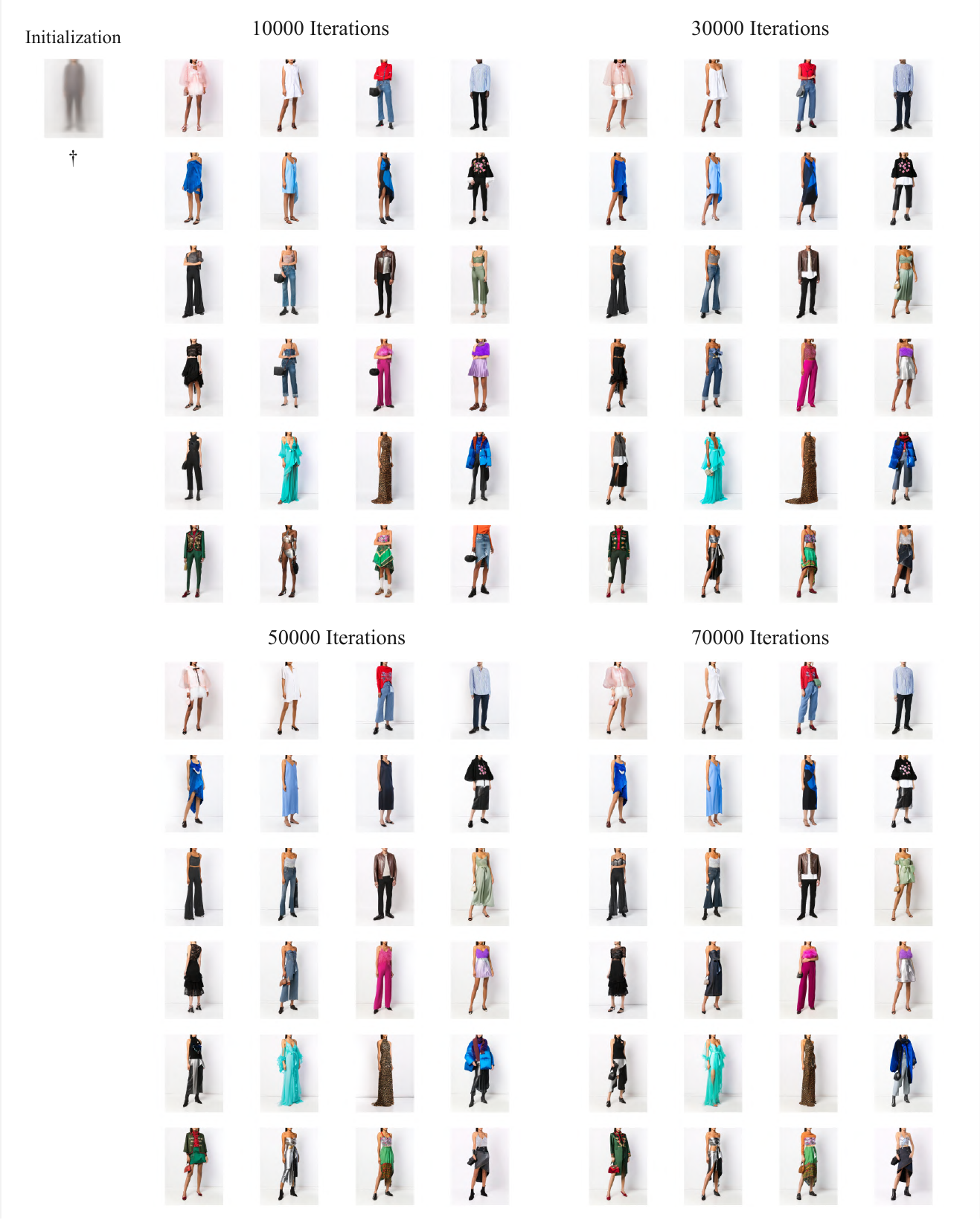}
    \caption{Comparison between training iterations on model generation for Mean Offset Training + Inference.}
    \label{fig:model_iteration}
\end{figure*}

The best garment outcomes are shown at $\alpha_T = 2.30 \times 10^{-4}$, but it still fails to consistently generate uniform white backgrounds (see top right blue mao collar shirt and bottom right patchwork skirt in Fig.~\ref{fig:garment_linear_end}) and model outputs are still unpredictable. Lowering $\alpha_T$ to $3.59 \times 10^{-5}$ and $5.67 \times 10^{-6}$ generates models that are more consistently centered in the image, but images for both garments and models are underexposed with substantial degradation in quality (compare with our Mean Offset Training 50k iteration results in Fig.~\ref{fig:garment_iteration} and Fig.~\ref{fig:model_iteration}). 

Small $\alpha$ values create a much harder denoising problem because the amount of noise added is significantly more. Additionally, we are scaling $\alpha_t$ to lower values for many timesteps $t$, and pre-trained sd v1.5~\cite{rombach2021highresolution} is not trained to handle large amounts of noise in these timesteps. This results in the need to re-train a greater number of weights and may require a larger dataset to avoid deterioration in image quality.

\begin{table}[]
    \centering
    \begin{tabular}{ |c|c| } 
     \hline
     Linear End Parameter & $\alpha_T$ \\ 
     \hline
     0.012 & 0.0046600  \\ 
     0.02 & 0.0002300 \\ 
     0.025 & 0.00003594 \\ 
     0.03 & 0.00000567 \\ 
     \hline
    \end{tabular}
    \caption{We show linear end values and each corresponding $\alpha_T$ and $\sqrt{\alpha_T}$. Standard training procedures use linear end $0.012$.}
    \label{tab:linear_end_alphas}
\end{table}

\section{PCA-K Inference on Stable Diffusion v1.5}
\begin{figure*}[htp!]
    \centering
    \includegraphics[width=\textwidth, trim=4 4 4 4,clip]{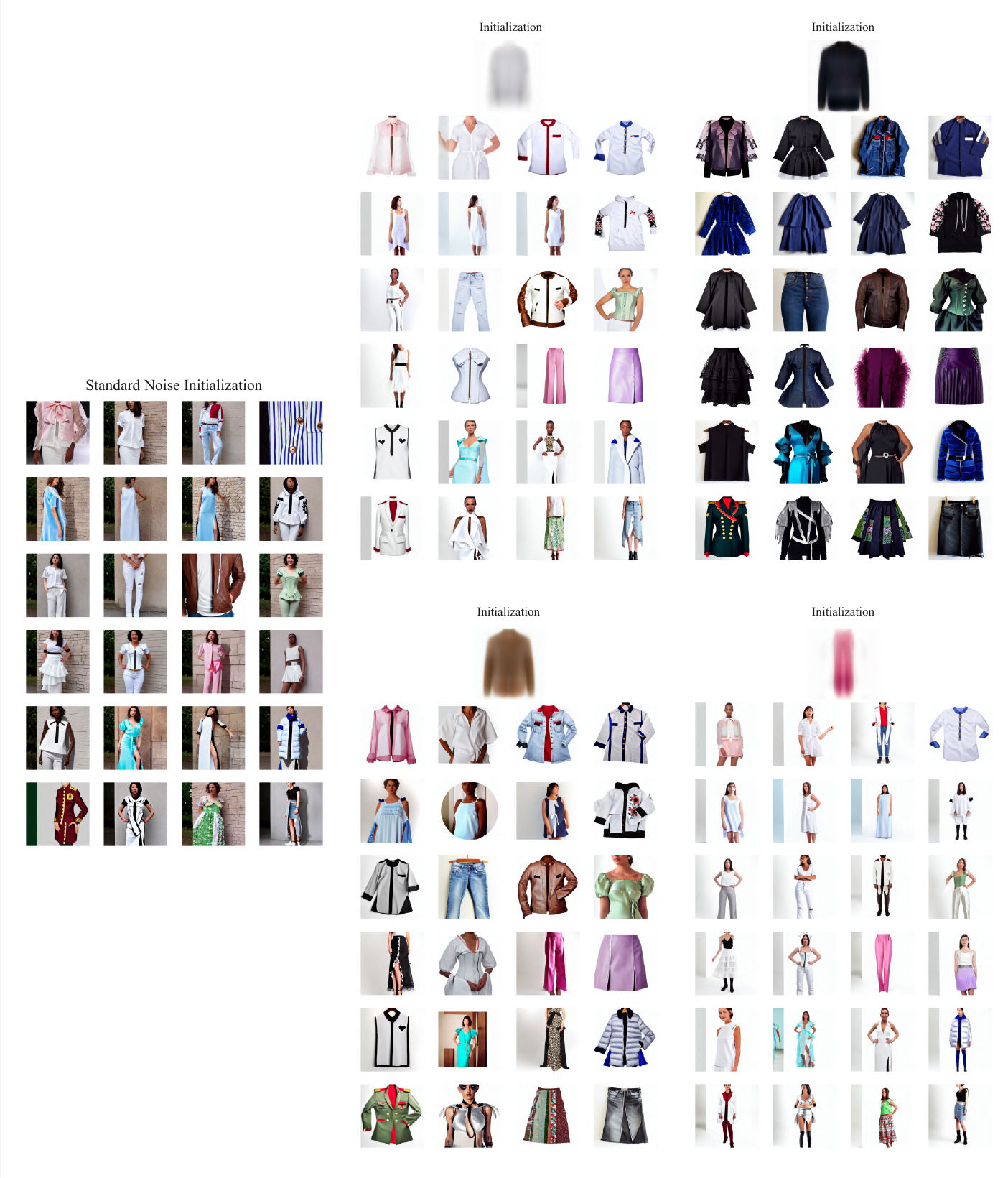}
    \caption{PCA-K Inference (K=3) results for garments on Stable Diffusion v1.5.}
    \label{fig:garment_inference}
\end{figure*}

\begin{figure*}[htp!]
    \centering
    \includegraphics[width=\textwidth, trim=4 4 4 4,clip]{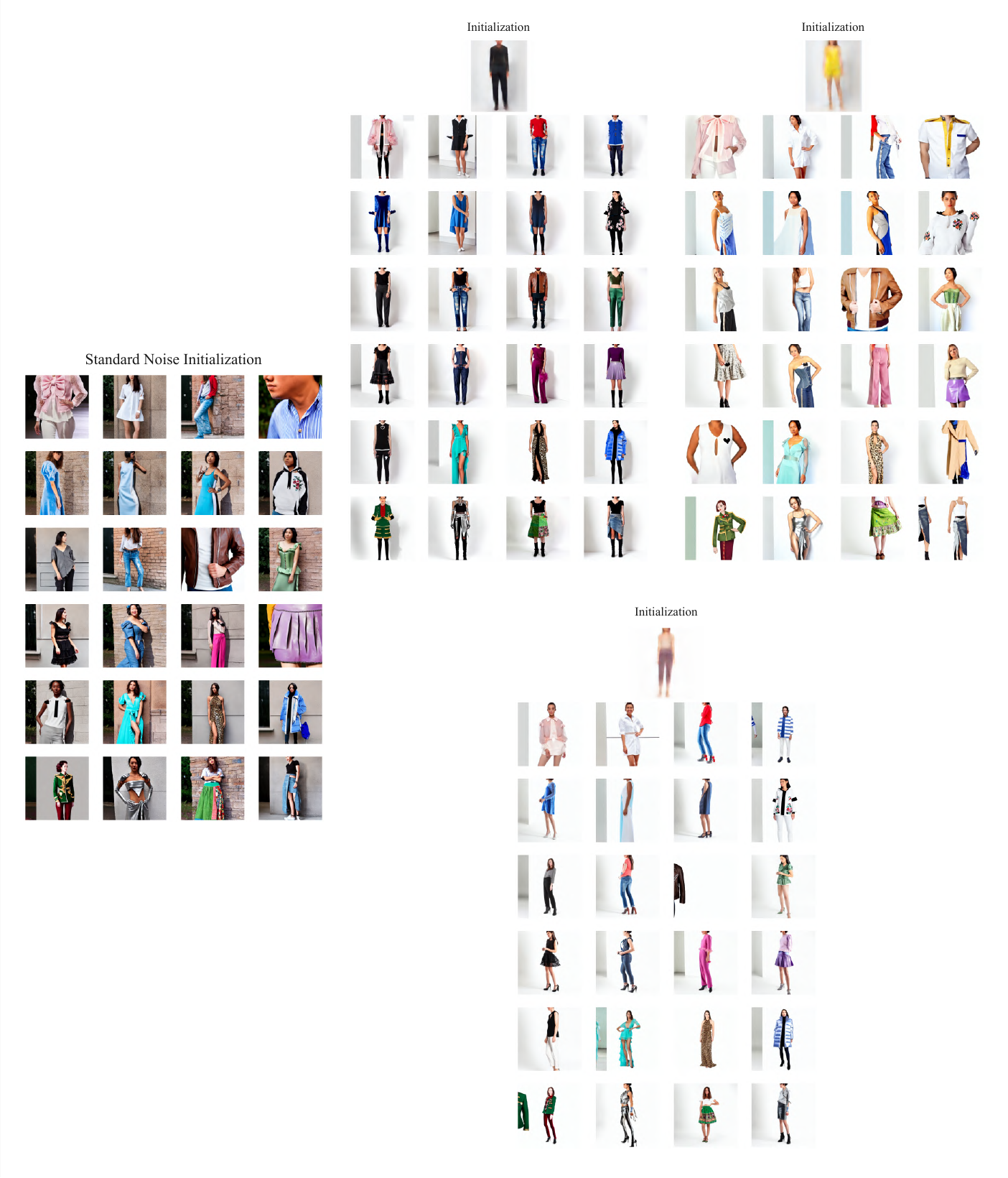}
    \caption{PCA-K Inference (K=3) results for models on Stable Diffusion v1.5.}
    \label{fig:model_inference}
\end{figure*}

Applying PCA-K Inference on pretrained Stable Diffusion v1.5 (sd v1.5)~\cite{rombach2021highresolution} can shift the outputted image distribution closer to our desired properties \emph{without} any fine-tuning. In Fig.~\ref{fig:garment_inference} and Fig.~\ref{fig:model_inference}, we show how sd v1.5's results change with PCA-3 Inference. With standard noise initialization, we see that none of the images for garments or models are acceptable retail images. However, using PCA-K Inference, the outputted image distribution is closer to satisfying our desired properties, and some may even be acceptable as retail images.

\section{Additional Materials for PCA-K Inference and PCA-K Training}
\subsection{PCA-K Inference Ablations}
\begin{figure*}[htp!]
    \centering
    \includegraphics[width=\textwidth, trim=4 4 4 4,clip]{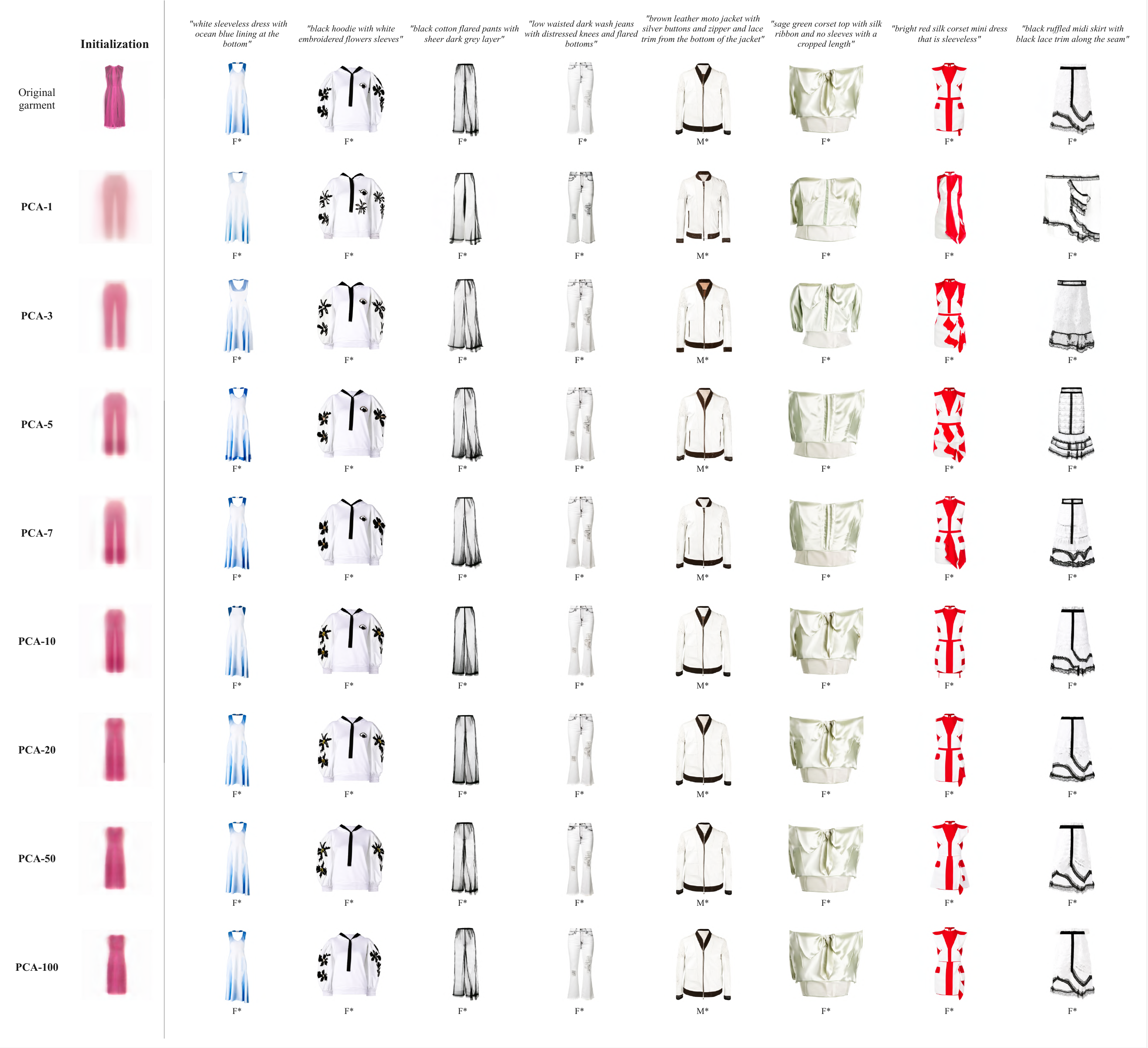}
    \caption{DDIM Training + PCA-K Inference at varying K values for a pink dress.}
    \label{fig:pca_comparison_1}
\end{figure*}

\begin{figure*}[htp!]
    \centering
    \includegraphics[width=\textwidth, trim=4 4 4 4,clip]{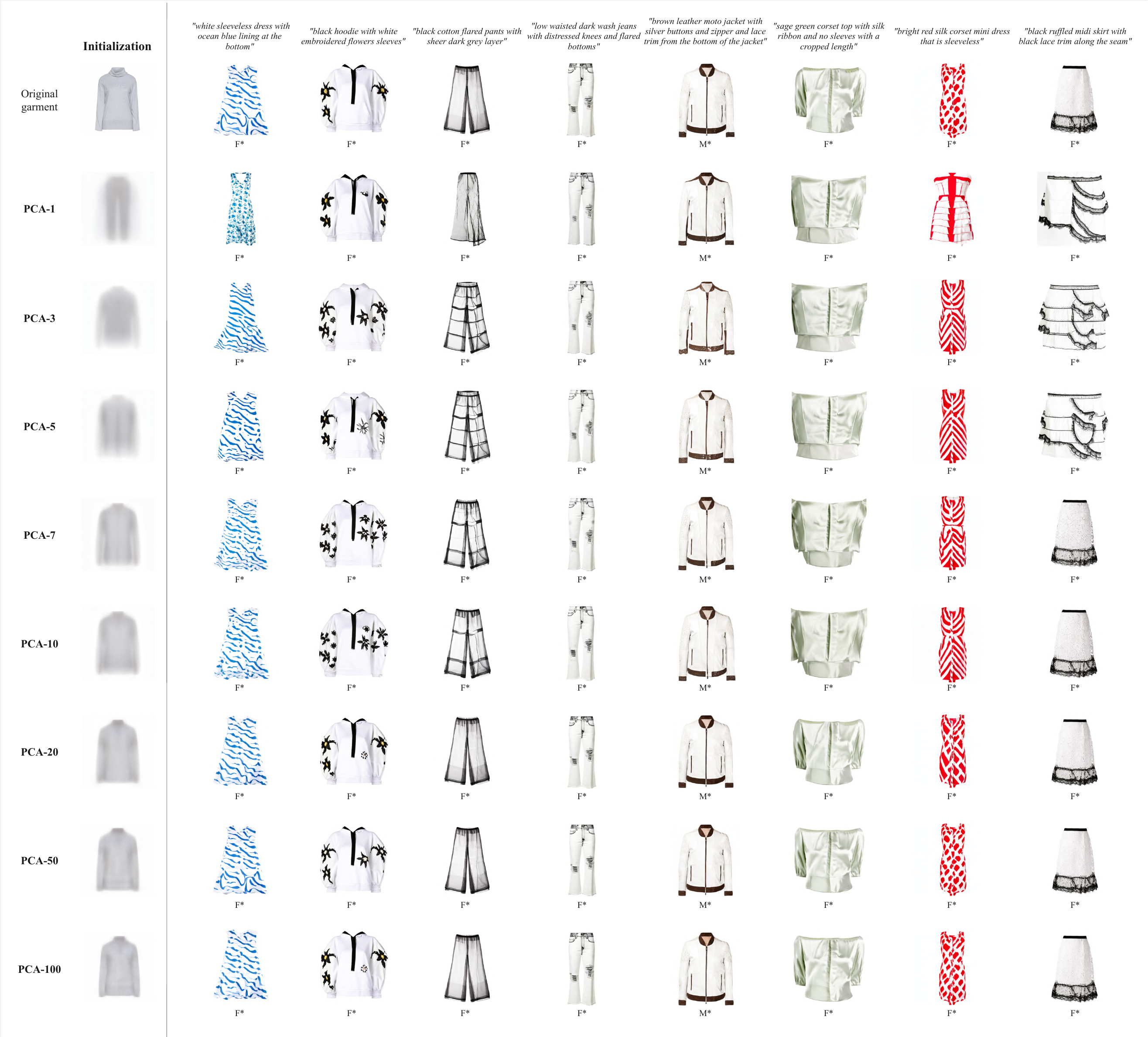}
    \caption{DDIM Training + PCA-K Inference at varying K values for a gray sweater.}
    \label{fig:pca_comparison_2}
\end{figure*}

\begin{figure*}[htp!]
    \centering
    \includegraphics[width=\textwidth, trim=4 4 4 4,clip]{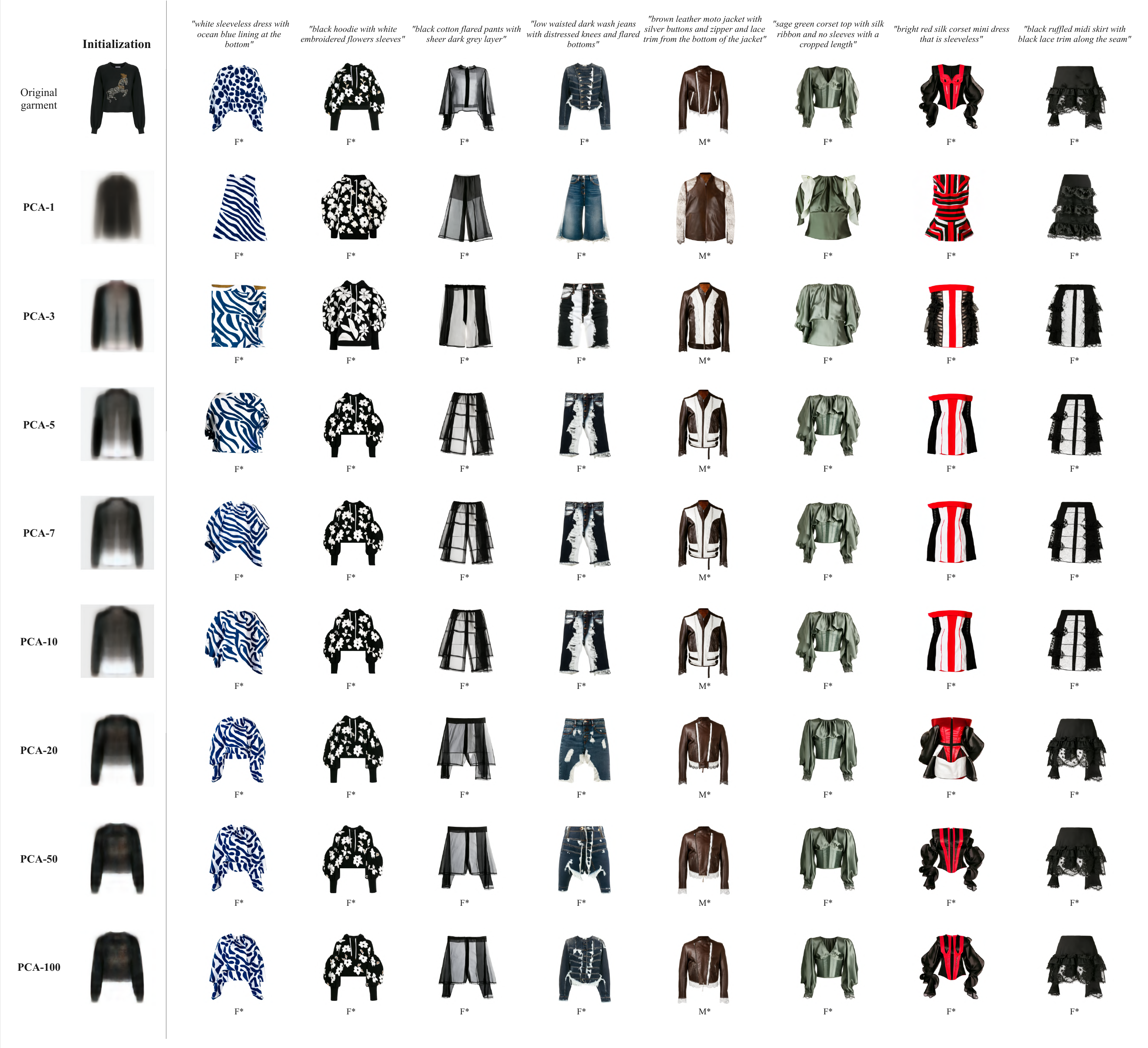}
    \caption{DDIM Training + PCA-K Inference at varying K values for a black long-sleeve.}
    \label{fig:pca_comparison_3}
\end{figure*}

In general, PCA is a reasonable and easy-to-sample approximation for $P(images)$ because the denoiser behaves similarly between a real image and a PCA-projected image. We show different garment images used as initialization during inference and how results change when these images are projected to different numbers of principal components. We use the standard fine-tuned model on our dataset from our main text (i.e., DDIM training). We use a pink dress (Fig.~\ref{fig:pca_comparison_1}), gray sweater (Fig.~\ref{fig:pca_comparison_2}), and a black long-sleeve (Fig.~\ref{fig:pca_comparison_3}) as reference. 

Lighter contrast garments (the pink dress and gray sweater in Fig.~\ref{fig:pca_comparison_1} and Fig.~\ref{fig:pca_comparison_2}) tend to behave similarly with as little as seven principal components. However, the black sweater (Fig.~\ref{fig:pca_comparison_3}) provides much higher contrast, and the denoiser tends to pick up more information from the initialization. Notice that the sleeves of the projected image are much darker than the center of the garment in PCA-3, PCA-5, PCA-7, and PCA-10. This causes generated results in columns 4, 5, 6, 8, and 9 to be darker in those sleeve regions and lighter in the center regions. Once we use a projection with a darker middle region (e.g., in PCA-20), the generated garments look similar to the original garment initialization result.

\subsection{Mean Offset Training}
We visualize Mean Offset Training results for all 24 text prompts and show our method can converge to a stable model within 50k iterations. Training parameters are identical to our fine-tuning procedure in the main text. In Fig.~\ref{fig:garment_iteration} and Fig.~\ref{fig:model_iteration}, we compare 10k, 30k, 50k, and 70k training iteration results. Notice that generated garment images satisfy all image properties (1)-(5) within 50k iterations and generated model images satisfy these properties even faster (within 30k iterations). 

\subsection{PCA-K Training For $K>0$}
\begin{figure*}[htp!]
    \centering
    \includegraphics[width=1.0\textwidth, trim=4 4 4 4,clip]{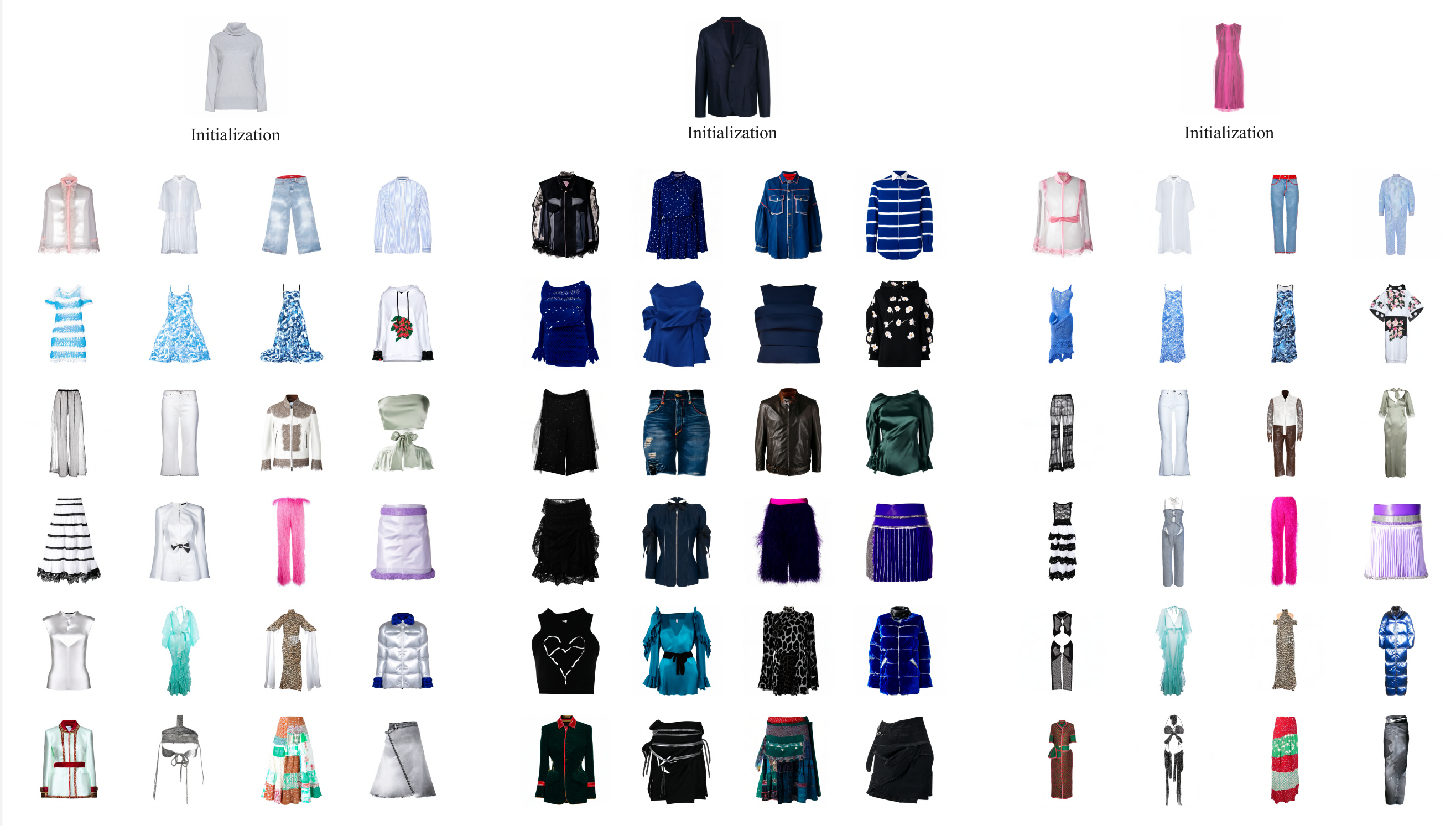}
    \caption{Comparison results between different initialization with standard training (DDIM Training + Inference).}
    \label{fig:initialization_difference_standard}
\end{figure*}

\begin{figure*}[htp!]
    \centering
    \includegraphics[width=1.0\textwidth, trim=4 4 4 4,clip]{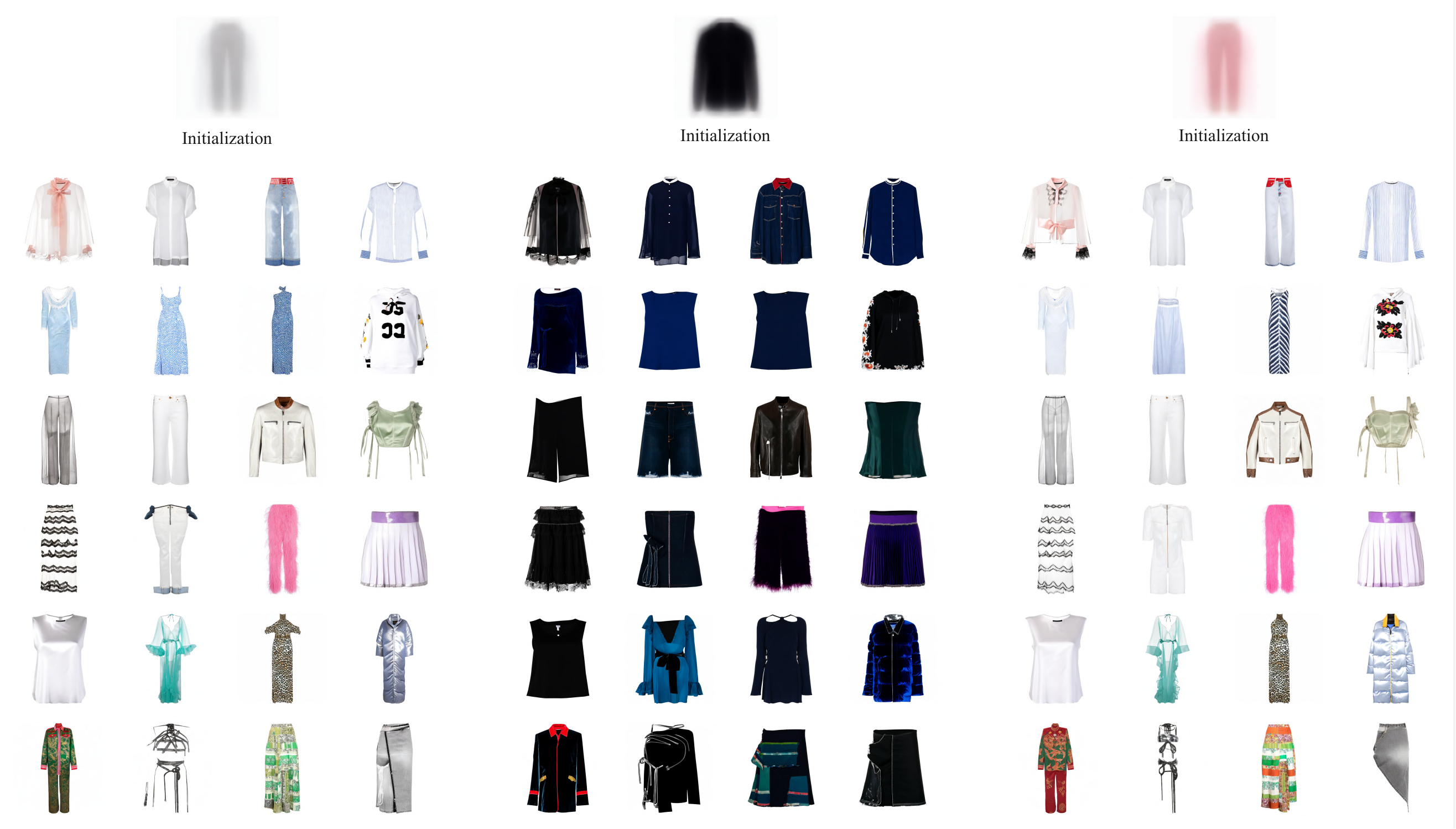}
    \caption{Comparison results between different initialization with PCA-1 Training + Inference.}
    \label{fig:initialization_difference_pca_1}
\end{figure*}

\begin{figure*}[htp!]
    \centering
    \includegraphics[width=1.0\textwidth, trim=4 4 4 4,clip]{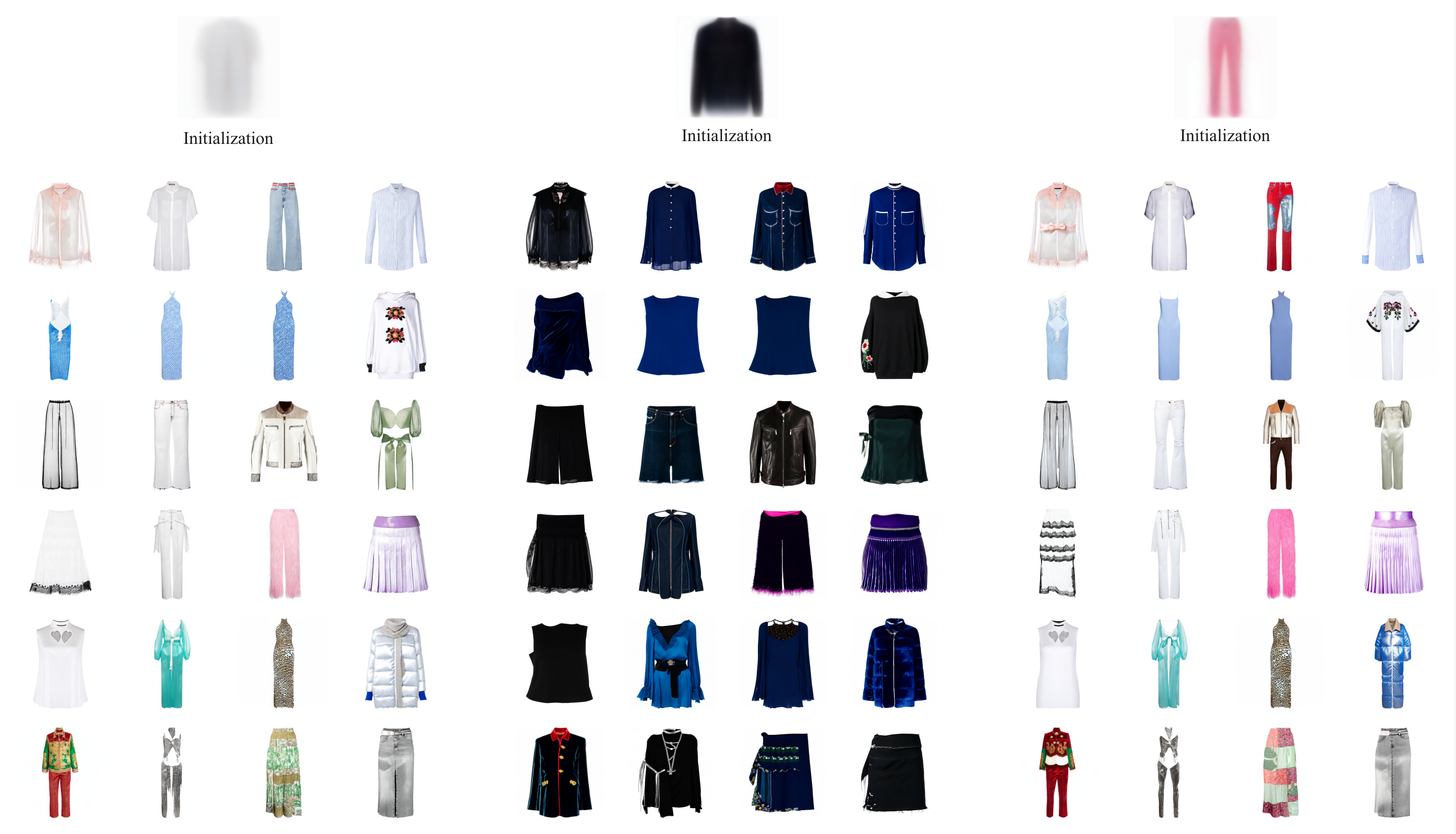}
    \caption{Comparison results between different initialization with PCA-3 Training + Inference.}
    \label{fig:initialization_difference_pca_3}
\end{figure*}

\begin{figure*}[htp!]
    \centering
    \includegraphics[width=1.0\textwidth, trim=4 4 4 4,clip]{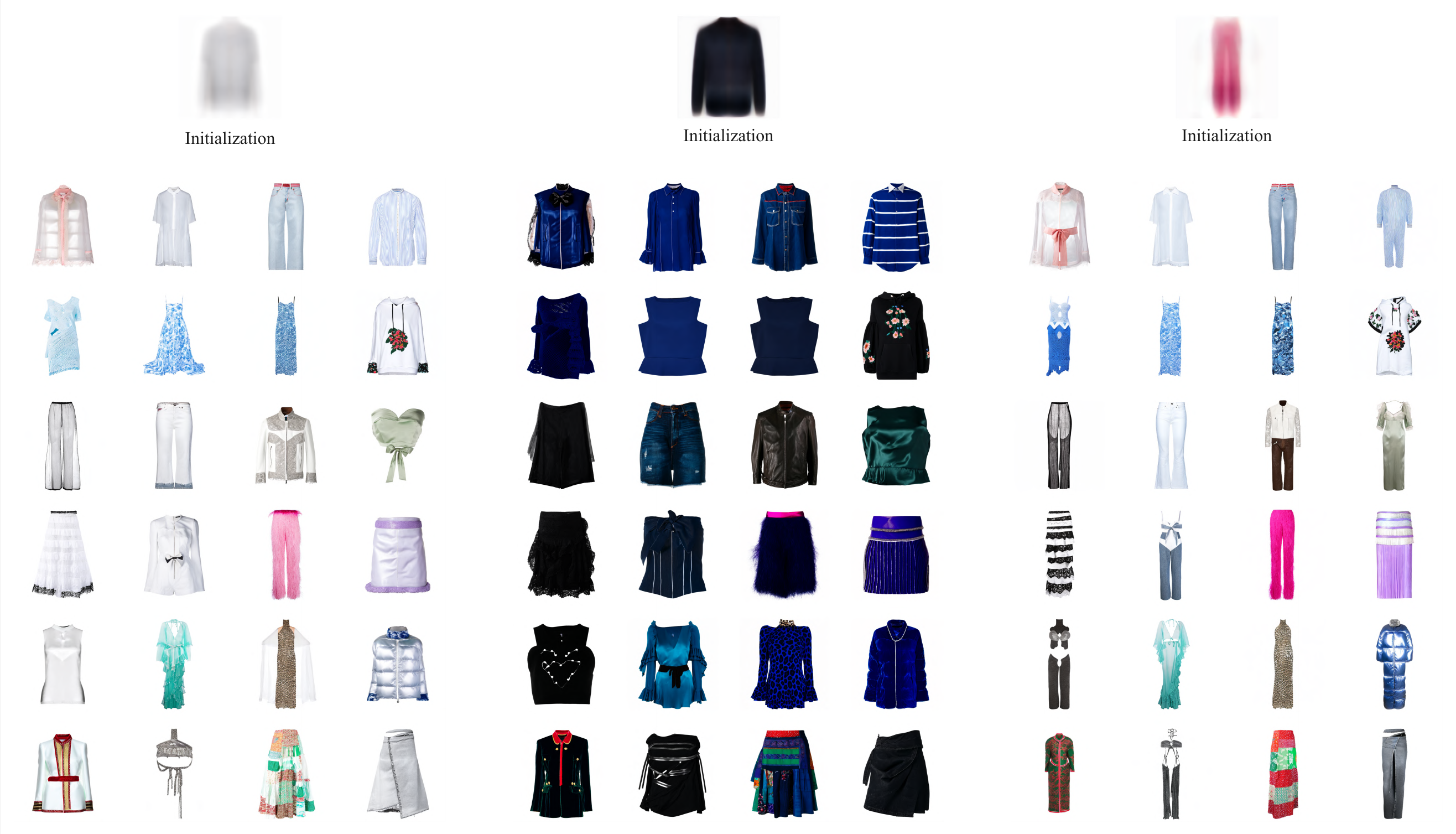}
    \caption{Comparison results between different initialization with PCA-10 Training + Inference.}
    \label{fig:initialization_difference_pca_10}
\end{figure*}

PCA-K Training for $K>0$ exhibits issues during inference due to a disconnect between the initialization and the text. In Fig.~\ref{fig:initialization_difference_standard}, we show results from standard fine-tuning and compare results to PCA-1 Training + Inference (Fig.~\ref{fig:initialization_difference_pca_1}), PCA-3 Training + Inference (Fig.~\ref{fig:initialization_difference_pca_3}), and PCA-10 Training + Inference (Fig.~\ref{fig:initialization_difference_pca_10}). For PCA-K Training + Inference ($K > 0$), the denoiser still utilizes color and shape information from the initialization and does not fully respect the text (e.g. black sweater initialization causes all the garments to be dark even if the text indicates lighter colors). We leave improving PCA-K ($K>0$) to future work. One suggestion is to use Canonical Correlation Analysis (CCA) to build relationships between PCA-K initializations and text features. This could help the network sample a smarter PCA-K initialization whose features correlate better with the text.




 \newpage
 {\small
 \bibliographystyle{ieee_fullname}
 \bibliography{supplement}
 }